\documentclass[10pt,twocolumn]{article} 
\usepackage{simpleConference}
\usepackage{times}
\usepackage{epsfig}
\usepackage{graphicx}
\usepackage{amsmath}
\usepackage{amssymb}
\usepackage{caption}
\usepackage{subcaption}
\usepackage{xcolor}
\usepackage{color,soul}
\usepackage{multirow}
\usepackage{booktabs}
\usepackage{enumitem}
\usepackage{wrapfig}
\usepackage{url,hyperref}
\begin{document}

\title{Multi-Class Multi-Instance Count Conditioned Adversarial Image Generation}

\author{Amrutha Saseendran\\
Bosch Center for Artificial Intelligence\\
{\tt\small Amrutha.Saseendran@de.bosch.com}
\and
Kathrin Skubch\\
Bosch Center for Artificial Intelligence\\
{\tt\small Kathrin.Skubch@de.bosch.com}
\and
Margret Keuper\\
University of Mannheim\\
{\tt\small keuper@uni-mannheim.de}
}

\maketitle
\thispagestyle{empty}

\begin{abstract}
Image generation has rapidly evolved in recent years. 
Modern architectures for adversarial training allow to generate even high resolution images with remarkable quality. 
At the same time, more and more effort is dedicated towards controlling the content of generated images. 
In this paper, we take one further step in this direction and propose a conditional generative adversarial network (GAN) that generates images with a defined number of objects from given classes. 
This entails two fundamental abilities (1) being able to generate high-quality images given a complex constraint and (2) being able to count object instances per class in a given image. 
Our proposed model modularly extends the successful StyleGAN2 architecture with a count-based conditioning as well as with a regression sub-network to count the number of generated objects per class during training.
In experiments on three different datasets, we show that the proposed model learns to generate images according to the given multiple-class count condition even in the presence of complex backgrounds.
In particular, we propose a new dataset, CityCount, which is derived from the Cityscapes street scenes dataset, to evaluate our approach in a challenging and practically relevant scenario. 
\end{abstract}
\section{Introduction}
\label{introduction}
\begin{figure}[h]
\centering
  \begin{tabular}{@{\hspace{.0cm}}c@{\hspace{.1cm}}c@{\hspace{.1cm}}c@{\hspace{.1cm}}c@{\hspace{.1cm}}}
 &\multicolumn{3}{c}{CityCount Examples}\\
 &\includegraphics[width=0.15\textwidth]{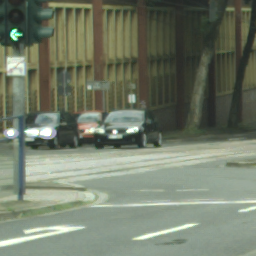}&
 \includegraphics[width=0.15\textwidth]{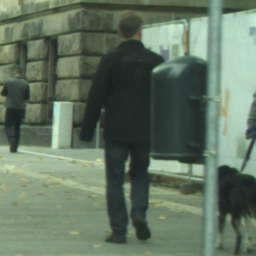} &
  \includegraphics[width=0.15\textwidth]{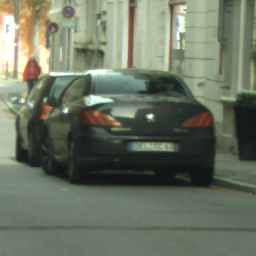}
  \\
  & \multicolumn{3}{c}{Generated Images}\\
 & \includegraphics[width=0.15\textwidth]{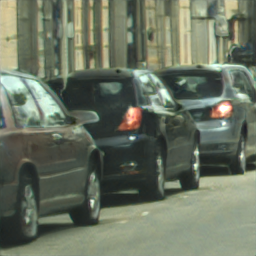}&
 \includegraphics[width=0.15\textwidth]{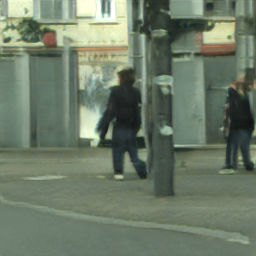} &
  \includegraphics[width=0.15\textwidth]{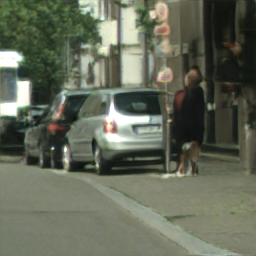}
 \\
&3 cars &2 persons &2 cars + 1 person\\
 \end{tabular}
\caption{Real and generated CityCount images by our model based on the multiple-class count input.} 
\label{fig:teaserpics}
\end{figure}
Developmental studies show that the human brain is endowed with a natural mechanism for understanding numerical quantities~\cite{numbersense,children1990}. Even young children have an abstract understanding of numeracy and can generalize the concept of counting from one category to another (\textit{e.g.} from objects to sounds)~\cite{children1990}.
While counting  
object instances is relatively easy for humans, it is challenging for deep learning and computer vision algorithms,
especially when objects from multiple classes, e.g. persons and cars,
are considered. 
In this paper, we take a step towards such elementary visual reasoning by addressing the generation of images conditioned on the number of object instances per object class.
We are particularly interested in the complex case where objects from \textit{multiple classes} are present in the same image (compare Figure~\ref{fig:teaserpics}).
This is a fundamental vision task, which can even be solved by small children~\cite{numbersense}, but remains an unsolved problem in computer vision. 
Apart from that, many practical applications can benefit from the capability to generate images respecting numerical constraints. 
Especially, it aids the generation of additional diverse training data for visual question answering and counting approaches.
Further, the generation of technical designs based on the number of different components is of particular interest in the field of topology design, where data-based approaches have recently been explored successfully in applications ranging from molecular design~\cite{NIPS2018_7978} for chemical applications to product design~\cite{10.1115/1.4044229} for aesthetics or engineering performance.

In this paper, we propose to solve \textit{multiple-class count} (MC$^2$) conditioned image generation (\textit{i.e.} the generation of images conditioned on the number of objects of different classes, that are visible in the image) as a modular extension to the state-of-the-art network for adversarial image generation, StyleGAN2~\cite{karras2019StyleGAN2}.
We further argue that object counting should be considered as a multi-class regression problem. While this approach is simple, it allows the similarity between neighboring numbers to be naturally encoded in the network and to transfer the ability to count from one class to another.
This will ideally make our network learn to generalize the concept of counting from one object class to another, meaning that it can for example see images of "two cars and one person" at training time and deduce the appearance of "two persons" at inference time.
To the best of our knowledge this is the first attempt to evaluate the potential of GANs to generate images based on the multiple object class count.

We validate the proposed approach in two lines of experiments. 
First, we evaluate the generative performance of our model on synthetic data generated according to the CLEVR~\cite{clevr} dataset as well as on real data from the SVHN~\cite{svhn} dataset.
We further propose a new, challenging real-world dataset, CityCount, which is derived from the well-known street scenes dataset Cityscapes~\cite{Cordts2016Cityscapes}.
The CityCount dataset comprises of various crops from Cityscapes images which contain specific numbers of objects from the important classes, \textit{car} and \textit{person}.
The dataset includes various challenging scenarios such as diverse and complex backgrounds, object occlusions, varying object scales and scene geometry.
Samples from the CityCount dataset and generated samples from our model are shown in    Figure \ref{fig:teaserpics}.
In the second line of experiments, we show that the images generated by MC$^2$-StyleGAN2 can be used to enhance the size and quality of training data for count prediction networks, trained on images from CLEVR and CityCount.

\section{Related Work}
\label{relwork}
\noindent\textbf{Generative adversarial networks (GANs)} - GANs~\cite{gan} have rapidly evolved to being the most promising trend for the generation of diverse photo-realistic images. 
Deep convolutional GAN (DCGAN)~\cite{dcgan} demonstrated the potential of convolutional neural networks in this context for the first time. 
A considerable amount of research was devoted to improve the training stability of GANs
~\cite{gulrajani2017wgan,karras2017ProGAN,miyato2018spectral_norm} and to develop more evolved architectures~\cite{brock2018BigGAN,karras2019StyleGAN,radford2015dcgan}. 
Conditioning GANs (CGAN) on explicit information was first introduced by Mirza \textit{et~al.}~\cite{cgan}.
Since then, various approaches have been proposed
to improve the controllability of GANs.
Many of these require extensive additional information
such as class labels and/or natural language descriptions, 
e.g. image captions for text-to-image or 
text-to-video generation~\cite{text2video,spatialgan,cgan,text2image}.
Other variants of conditioning GANs include auxiliary classifier GAN (ACGAN)~\cite{acgan}, twin auxiliary classifier GAN (TACGAN)~\cite{tacgan} and projection based conditioning methods~\cite{projectioncgan}.
ACGAN extends the loss function of GAN with an auxiliary classifier to generate images. TACGAN further improves the divergence between real and generated data distribution of ACGAN by an additional network that interacts with both generator and discriminator.
In projection based methods, the condition is projected to the output of the discriminator by considering the inner product of the conditional variable and the feature vector of images.
SpatialGAN~\cite{spatialgan} propose a method for multiple conditioning
with bounding box annotations and class labels of objects, and image captions 
to control the image layout in terms of object identity, size, position and number. In their method, object bounding boxes are provided at test time so the idea of count does not need to be learned. 
In~\cite{vaeunet} the authors propose a variational U-Net architecture to condition the image generation on shape or appearance.
Various approaches have also been suggested 
to control the image generation process of GANs
in applications such as image-to-image-translation~\cite{pix2pixgan,cyclegan} or attribute transfer~\cite{AttGan,liu2019stgan}.
Our work is related to ACGAN, with focus on the problem of multiple-class
counting using regression.

Based on the high-resolution architecture introduced in~\cite{karras2017ProGAN}, StyleGAN~\cite{karras2019StyleGAN} employs adaptive instance normalization~\cite{huang2017AdaIN} based feature map re-weighting
to facilitate the manipulation of images over multiple latent spaces, encoding different style properties.
 StyleGAN2~\cite{karras2019StyleGAN2} improves over StyleGAN and avoids some characteristic generation artifacts.
Recently, a new technique was proposed~\cite{Karras2020ada} to achieve state of the art results with StyleGAN2 even when the training data is limited.
While these approaches allow implicit conditioning of image contents for example on given styles, they do not enable to steer explicit properties of a generated image such as the number of generated object instances per object class. 
Our proposed model introduces an extension to StyleGAN2, that facilitates such an explicit conditioning.

\noindent\textbf{Counting approaches} - One way to count objects in an image is to first localize and classify them using an object detection network and then count all found instances. 
While this approach is effective, it also requires additional class labeled bounding box or object prototype information~\cite{segmentcount,cellcountdetection,detectioncount}. 
Adapting these approaches for conditional image generation will require additional information such as pre-defined locations of the objects of interest during training.
Other methods rely on recurrent neural network architectures and attention mechanisms~\cite{countrnn2,countrnn,resnest}. Thus, they can not easily be applied in our problem setting.
Density estimation based counting methods~\cite{densitycount} show that learning to count can be achieved without prior detection and are more reliable in severe occlusion scenarios.
Multiple approaches have been proposed to counting object instances in images, for example in the context of visual question answering~\cite{vqa,qacount,babi}. 
In~\cite{casualvqa}, Agarwal \textit{et~al.} suggest to generate training data for this task by modifying the number of objects using cropping and inpainting.

In this paper, we attempt to guide the image generation process solely by conditioning on the 
number of objects of pre-defined classes in the images, while a reasonable spatial layout is to be inferred from the training data distribution.
Instead of addressing single object class counting as seen in~\cite{deeplearningcount,nalu}, where convolutional or recurrent neural networks are used to count digit occurrences, our approach focuses on counting object instances from multiple classes during \textit{generation}.
We introduce an extension to the StyleGAN2 architecture by integrating an additional regression network to the discriminator to facilitate image generation based on the number of objects per class.
Based on the findings in~\cite{Chen2018EfficientAA}, our network employs dense blocks in the generator architecture to ease the propagation of the count constraint as well as the regression loss of the count network.
\section{Multiple Class Count Conditioned Image Generation}
\label{method}
In this section, we introduce the proposed extension to StyleGAN2 for multiple-class count based image generation, MC$^2$-StyleGAN2.
\begin{figure}
         \includegraphics[width=1.1\columnwidth]{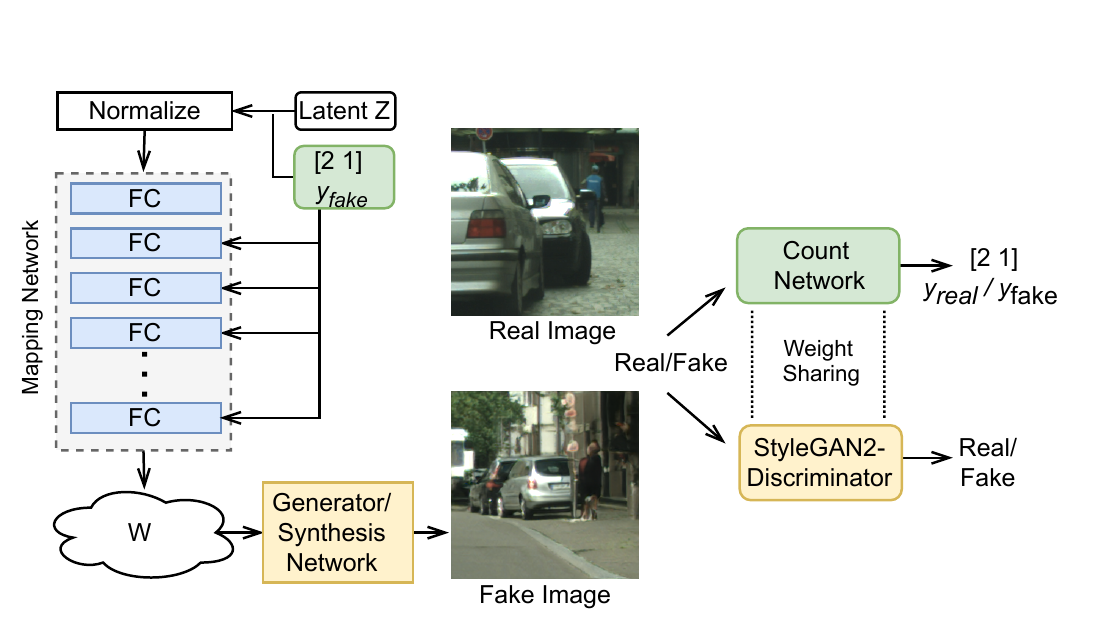}
     \caption{MC$^2$-StyleGAN2 architecture: 
     The input to the generator is a multiple-class count vector where each index of the vector corresponds to each object class and the value at each index represents the multiplicity of the corresponding object class.
     In the given CityCount example, the count vector [2,1] corresponds to 2 cars and 1 person respectively.}
     \label{MCCstyleGAN2}
\end{figure}
\subsection{MC$^2$-StyleGAN2} 
We borrow the architectural specifications of the generator and discriminator from StyleGAN2 and extend the model for our application.
The input to the generator is a multiple-class count vector, where each index of the vector corresponds to a different object class and the value at each index represents the number of objects from the corresponding object class.
The generative part of our model includes a mapping network to map the combination of latent vector and the count constraint to an intermediate latent vector $w$ and a generator/synthesis network to generate images as shown in Figure~\ref{MCCstyleGAN2}.
To the first layer of the mapping network, we provide a combination of randomly sampled noise and our multiple-class count vector, that specifies which objects and how many of each of them are required in the output image.
The count vector is also concatenated to every layer in the mapping network as shown in Figure~\ref{MCCstyleGAN2}.
In the generator network, we introduce dense like skip connections where the output from each block is connected to its succeeding blocks.
As shown in Figure~\ref{MCCstyleGAN2}, the real/generated images are passed through two pathways, (1) an adversarial pathway to classify the input images as real/fake and (2) a count regression pathway, to predict the object class and their multiplicity in the input image.
The weight sharing between the two sub-networks regularizes the discriminator and reduces the memory consumption during training.
\subsection{Adversarial Training with Count Loss} 
The generator $G$, uses both the latent noise distribution $z \sim \mathcal{N}(0,1)$ and a multiple-class count vector $\textbf{c}=[c_1, c_2, \ldots, c_n]$ that represents $n$ different object classes and their respective multiplicity $c_i, i=,\ldots, n$,
to generate fake images $x_{\mathrm{fake}} = G(z,\mathbf{c})$.
The discriminator $D$ aims to distinguish between these fake images and real images $x_{\mathrm{real}}$.
We denote the data distribution as $x \sim p_{data}(x)$.
The additional count sub-network $C$ is trained to predict the per-class object count, $y_{\mathrm{fake}}$ for fake images and 
$y_{\mathrm{real}}$ for real images.
The adversarial objective of the network is expressed as
\begin{align}
&\mathcal{L}_{GAN}(G,D)&=\hspace{0.1cm}&\mathbb{E}_{\boldsymbol{x}\sim p_{\text{data}}(\boldsymbol{x})}[\log{D(\boldsymbol{x})}]\hspace{0.1cm}+\nonumber\\
&&&\mathbb{E}_{\boldsymbol{z}\sim p_{\boldsymbol{z}}(\boldsymbol{z})}[\log{(1-D(G(\boldsymbol{z|c})))}].
\end{align}
The multiple-class count loss $\mathcal{L}_{MC^2}$
is defined as the euclidean distance between the predicted count $y_{\mathrm{real}}=C(x_{\mathrm{real}})$ and true count $\mathbf{c}$ of the real images, and the distance between the predicted count $y_{\mathrm{fake}}=C(x_{\mathrm{fake}})$ and the value of the count condition for the generated images. 
\begin{equation}
\mathcal{L}_{MC^2}(C) = {||C(x) - \mathbf{c}||}_2.
\label{eq:countloss}
\end{equation}   
The count loss thus enforces the generator to generate images with the desired number of object instances.

Hence, the total loss of the network is a combination of adversarial loss to match the distribution of real images with fake images and a count loss to enforce the network to generate images based on the specified input count.
The overall objective function of our method is,
\begin{equation}
\mathcal{L}_{MC^2-StyleGAN2}(G,D) = \mathcal{L}_{GAN}(G,D) + \lambda\mathcal{L}_{MC^2}(C),
\label{eq:countloss}
\end{equation}
where $\lambda$ steers the importance of the count objective. \\\\
\noindent\textbf{Implementation Details}
The models are trained with images of size $64\times 64$ for SVHN, $128\times 128$ for CLEVR images and $256\times 256$ for CityCount images.
All hyperparameters used are provided in the Appendix.
\section{Experimental Analysis}
\label{exp}
\begin{table}
\centering
\begin{minipage}{\textwidth} 
  \begin{tabular}{|c|cc|cc|}
    \hline
    Dataset     & \multicolumn{2}{c|}{Unconditioned}  &\multicolumn{2}{c|}{MC$^2$-StyleGAN2}  \\ 
    &\multicolumn{2}{c|}{StyleGAN2} &\multicolumn{2}{c|}{(Ours)} \\
    \cline{2-5}
    &MSE($\downarrow$)
    &FID($\downarrow$)
    &MSE($\downarrow$)
    &FID($\downarrow$)
 \\
    \hline\hline
    CLEVR-$2$  & $-$  &$7.241$   &\bf{0.004}  &\bf{7.979} \\
    CLEVR-$3$  &$-$  &$9.034$   &\bf{0.022}  &\bf{8.936} \\
    SVHN-$2$   &$-$  &$14.861$    &\bf{0.015}  &\bf{10.901}  \\
    CityCount$^*$
    & $-$    &$10.567$  &\bf{0.356}  &\bf{8.328}\\
    \hline
     \end{tabular}%
    \end{minipage}
\caption{Quantitative analysis across datasets - StyleGAN2 models. $^*$For CityCount we used StyleGAN2 with adaptive discriminator augmentation.~\cite{Karras2020ada}\label{tab_StyleGAN2}
}  
\end{table}
In the following, we evaluate our model in three different settings.
We quantitatively evaluate (1) the ability of the model to predict the multiple-class count in terms of Mean Squared Error (MSE) and (2) the quality of the images generated based on the learned count in terms of the Fr\'echet Inception Distance (FID).
The quantitative results for the three datasets are given in Table~\ref{tab_StyleGAN2}.
\begin{figure*}
 \scriptsize
 \selectfont
 \centering
 \subcaptionbox{CLEVR-$2$ and CLEVR-$3$(simple) - Count vector corresponds to number of cylinders, spheres and cubes. \label{fig:clevrs_generatedstylegan}}{
    \begin{tabular}{@{\hspace{.0cm}}c@{\hspace{.045cm}}c@{\hspace{.035cm}}c@{\hspace{.1cm}}c@{\hspace{.045cm}}c@{\hspace{.035cm}}c@{\hspace{.045cm}}c@{\hspace{.035cm}}c@{\hspace{.075cm}}c@{\hspace{.0cm}}
    }
    
\includegraphics[width=0.101\textwidth]{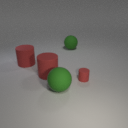} &  
\includegraphics[width=0.101\textwidth]{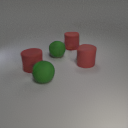} & 
\includegraphics[width=0.101\textwidth]{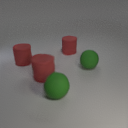} & 
\includegraphics[width=0.101\textwidth]{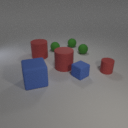} &  
\includegraphics[width=0.101\textwidth]{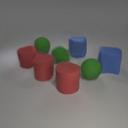} &
\includegraphics[width=0.101\textwidth]{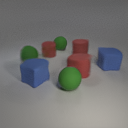} &
\includegraphics[width=0.101\textwidth]{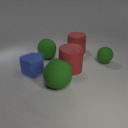} & 
\includegraphics[width=0.101\textwidth]{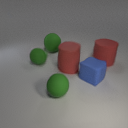} 
 \\
 real &\multicolumn{2}{@{\hspace{.01cm}}c@{\hspace{.01cm}}}{{[}3 2]} &real &\multicolumn{2}{@{\hspace{.01cm}}c@{\hspace{.01cm}}}{{[}3 3 2]}  &\multicolumn{2}{@{\hspace{.01cm}}c@{\hspace{.01cm}}}{{[}2 3 1]} \\
\end{tabular}}
  \subcaptionbox{CLEVR-$2$ - Count vector corresponds to number of cylinders and spheres. \label{fig:clevr_generatedstylegan}}{
    \begin{tabular}{@{\hspace{.0cm}}c@{\hspace{.035cm}}c@{\hspace{.015cm}}c@{\hspace{.035cm}}c@{\hspace{.015cm}}c@{\hspace{.035cm}}c@{\hspace{.015cm}}c@{\hspace{.0cm}}
    }
 
\includegraphics[width=0.12\textwidth]{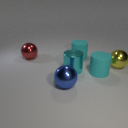} & 
\includegraphics[width=0.12\textwidth]{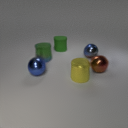} & 
\includegraphics[width=0.12\textwidth]{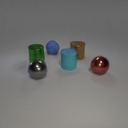} & 
\includegraphics[width=0.12\textwidth]{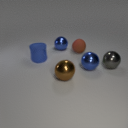} & 
\includegraphics[width=0.12\textwidth]{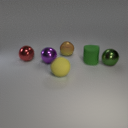} & 
\includegraphics[width=0.12\textwidth]{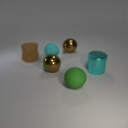} & 
\includegraphics[width=0.12\textwidth]{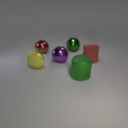} 

 \\
 real &\multicolumn{2}{@{\hspace{.01cm}}c@{\hspace{.01cm}}}{{[}3 3]} &\multicolumn{2}{@{\hspace{.01cm}}c@{\hspace{.01cm}}}{{[}1 5]} &\multicolumn{2}{@{\hspace{.01cm}}c@{\hspace{.01cm}}}{{[}2 4]} \\ 
\end{tabular}}
\subcaptionbox{CLEVR-$3$ - Count vector corresponds to number of cylinders, spheres and cubes. \label{fig:clevr3_generatedstylegan}}{
    \begin{tabular}{@{\hspace{.0cm}}c@{\hspace{.035cm}}c@{\hspace{.015cm}}c@{\hspace{.035cm}}c@{\hspace{.015cm}}c@{\hspace{.035cm}}c@{\hspace{.015cm}}c@{\hspace{.0cm}}
    }
\includegraphics[width=0.12\textwidth]{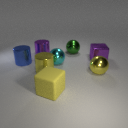} & 
\includegraphics[width=0.12\textwidth]{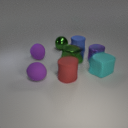} & 
\includegraphics[width=0.12\textwidth]{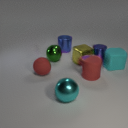} &
\includegraphics[width=0.12\textwidth]{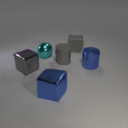} & 
\includegraphics[width=0.12\textwidth]{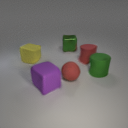} &
\includegraphics[width=0.12\textwidth]{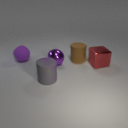} & 
\includegraphics[width=0.12\textwidth]{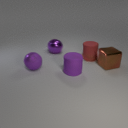} 
 \\
 real &\multicolumn{2}{@{\hspace{.01cm}}c@{\hspace{.01cm}}}{{[}3 3 2]}  &\multicolumn{2}{@{\hspace{.01cm}}c@{\hspace{.01cm}}}{{[}2 1 3]}  &\multicolumn{2}{@{\hspace{.01cm}}c@{\hspace{.01cm}}}{{[}2 2 1]} \\
 \end{tabular}}
\subcaptionbox{SVHN-$2$ - Count vector corresponds to per digit count. \label{fig:svhn_generatedstylegan}}{
    \begin{tabular}{@{\hspace{.0cm}}c@{\hspace{.045cm}}c@{\hspace{.025cm}}c@{\hspace{.045cm}}c@{\hspace{.025cm}}c@{\hspace{.045cm}}c@{\hspace{.025cm}}c@{\hspace{.045cm}}c@{\hspace{.025cm}}c@{\hspace{.0cm}}
    }
\includegraphics[width=0.092\textwidth]{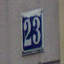} &
\includegraphics[width=0.092\textwidth]{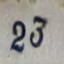} & 
\includegraphics[width=0.092\textwidth]{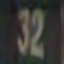} & 
\includegraphics[width=0.092\textwidth]{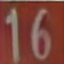} & 
\includegraphics[width=0.092\textwidth]{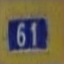} & 
\includegraphics[width=0.092\textwidth]{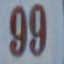} & 
\includegraphics[width=0.092\textwidth]{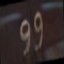} &
\includegraphics[width=0.092\textwidth]{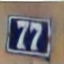} & 
\includegraphics[width=0.092\textwidth]{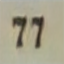}

 \\
 real &\multicolumn{2}{@{\hspace{.01cm}}c@{\hspace{.01cm}}}{{[}0 0 1 1 0 0 0 0 0 0]} &\multicolumn{2}{@{\hspace{.01cm}}c@{\hspace{.01cm}}}{{[}0 1 0 0 0 0 1 0 0 0]} 
 &\multicolumn{2}{@{\hspace{.01cm}}c@{\hspace{.01cm}}}{{[}0 0 0 0 0 0 0 0 0 2]} 
  &\multicolumn{2}{@{\hspace{.01cm}}c@{\hspace{.01cm}}}{{[}0 0 0 0 0 0 0 2 0 0]} 
  \\
 \end{tabular}}
\subcaptionbox{CityCount - Count vector corresponds to number of cars and persons. Boxes are drawn around objects of interest for ease of visualization. \label{fig:citycount_generatedstylegan}}{
    \begin{tabular}{@{\hspace{.0cm}}c@{\hspace{.03cm}}c@{\hspace{.03cm}}c@{\hspace{.03cm}}c@{\hspace{.03cm}}c@{\hspace{.03cm}}c@{\hspace{.0cm}}
    }
\includegraphics[width=0.15\textwidth]{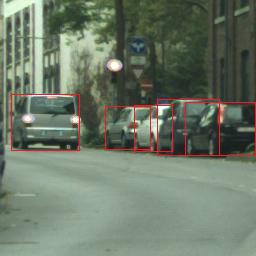} & 
\includegraphics[width=0.15\textwidth]{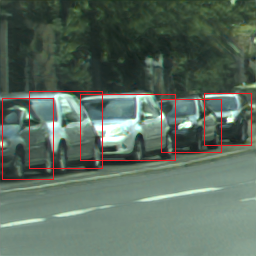} & 
\includegraphics[width=0.15\textwidth]{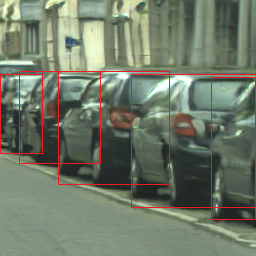} & 
\includegraphics[width=0.15\textwidth]{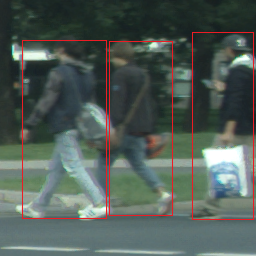} & 
\includegraphics[width=0.15\textwidth]{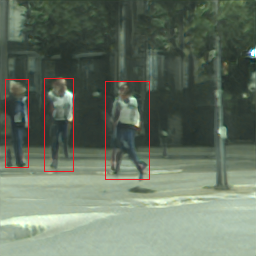} & 
\includegraphics[width=0.15\textwidth]{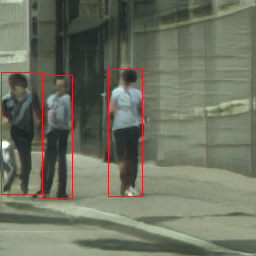} \\
 real &\multicolumn{2}{@{\hspace{.01cm}}c@{\hspace{.01cm}}}{{[}5 0]} & real  &\multicolumn{2}{@{\hspace{.01cm}}c@{\hspace{.01cm}}}{{[}0 3]} \\
\includegraphics[width=0.15\textwidth]{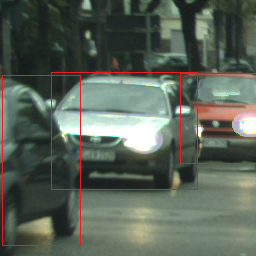} & 
\includegraphics[width=0.15\textwidth]{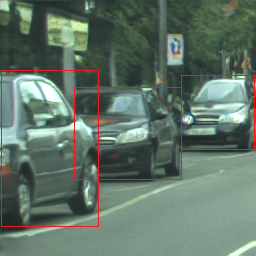} & 
\includegraphics[width=0.15\textwidth]{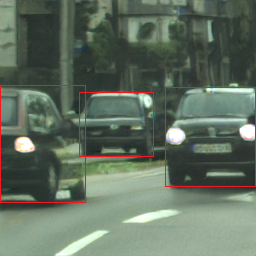} &
\includegraphics[width=0.15\textwidth]{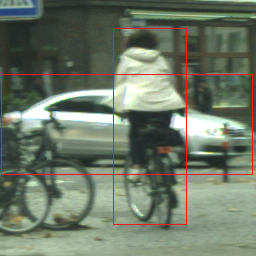} & 
\includegraphics[width=0.15\textwidth]{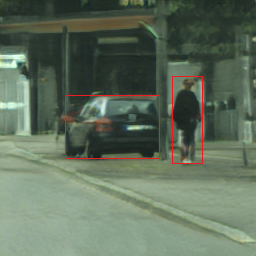} & 
\includegraphics[width=0.15\textwidth]{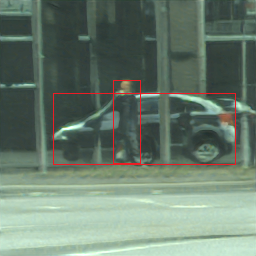} 
 \\
real &\multicolumn{2}{@{\hspace{.01cm}}c@{\hspace{.01cm}}}{{[}3 0]} & real  &\multicolumn{2}{@{\hspace{.01cm}}c@{\hspace{.01cm}}}{{[}1 1]} \\
\includegraphics[width=0.15\textwidth]{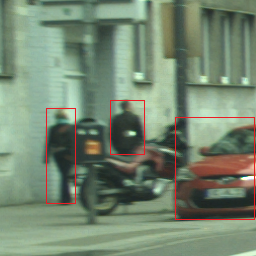} & 
\includegraphics[width=0.15\textwidth]{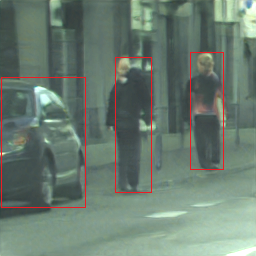} & 
\includegraphics[width=0.15\textwidth]{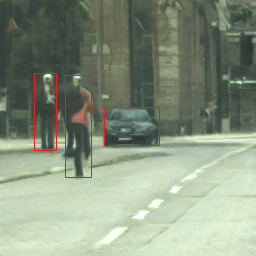} & 
\includegraphics[width=0.15\textwidth]{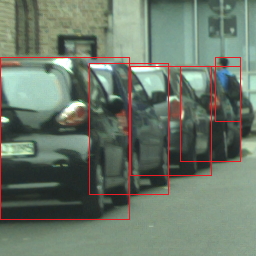} & 
\includegraphics[width=0.15\textwidth]{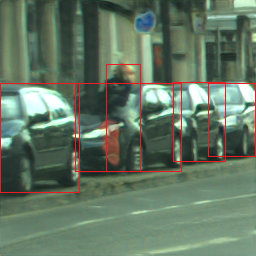} & 
\includegraphics[width=0.15\textwidth]{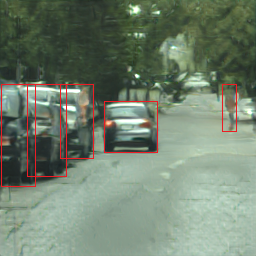} 
 \\
real &\multicolumn{2}{@{\hspace{.01cm}}c@{\hspace{.01cm}}}{{[}1 2]} & real  &\multicolumn{2}{@{\hspace{.01cm}}c@{\hspace{.01cm}}}{{[}4 1]} \\
\end{tabular}}
 \caption{Generated MC$^2$-StyleGAN2 images for different count combination across datasets.}
 \label{fig:MCCStylegan2}
 \end{figure*}
\subsection{CLEVR} 
The objective of the following experiment is to analyze the ability of the model to generate complex $3$D objects and layouts.
The well-known CLEVR dataset comprises images of different $3$D shapes, cylinders, cubes and spheres of varying colours.
For our experiments, we generate a total of $2000$ images for each count combination based on the implementation of CLEVR dataset~\cite{clevr}.
We consider two variants of CLEVR images, 
(1) CLEVR-$2$ with two shapes, cylinder and sphere, and at most six instances of each shape per image and      
(2) CLEVR-$3$  with three shapes, cylinder, sphere and cube, and at most three instances of each shape per image.
For our first line of experiments, we consider a simple setting, where we restrict shapes of the same class to be of the same color (red cylinders, green spheres and blue cubes).
The generated images shown in Figure~\ref{fig:clevrs_generatedstylegan} show that the proposed model learns to generate images based on the learned object count.
For further evaluation, we extend the experimental setting and consider CLEVR shapes with varying colors.
As shown in Figure~\ref{fig:clevr_generatedstylegan} and ~\ref{fig:clevr3_generatedstylegan}, the model captures the correlation of the count information even in a more complex setting, where the shape colors do not provide additional information.
It can also be observed that the model learned to place objects spatially in reasonable locations although no object bounding box annotations are provided.

Additionally, for count prediction analysis, we consider the performance of the count sub-network in the model.
We observed an average count accuracy of $96\%$ for CLEVR-$2$ and $92\%$ for CLEVR-$3$ 
 (a more detailed analysis of the count prediction on CLEVR-$2$ and CLEVR-$3$ is provided in the Appendix).
For CLEVR-$3$, the observed count prediction accuracy is comparatively lower than for CLEVR-$2$, potentially for two reasons, (1) the image distribution is highly complex due to the high number of objects in the image (maximum of nine objects per image) and (2) objects in the images are often overlapping significantly. \\ \\
\noindent\textbf{Interpolation and Extrapolation}
\begin{table}
\centering
\begin{minipage}{\textwidth} 
  \begin{tabular}{|c|c|c|c|}
    \hline
    \multirow{3}{*}{Dataset}  &\multicolumn{3}{c|}{MSE($\downarrow$)} \\
    \cline{2-4}
     &Seen &Unseen  &Unseen  \\
     & &interpolation &extrapolation
 \\
    \hline\hline
    CLEVR-$2$   &$0.004$  &$0.012$   &$0.084$ \\
    CLEVR-$3$ & $0.022$  &$0.029$   &$0.079$  \\
    \hline
     \end{tabular}%
    \end{minipage}
\caption{The Mean Squared Error (MSE) values observed for interpolation and extrapolation on unseen count combination from one object class to another in CLEVR images.\label{fig:interpolationclevr}
}  
\end{table}
We further examine the ability of the model to interpolate between count combinations and to extrapolate to unseen count combinations from one object class to another.
For interpolation
experiments, we train our model on a subset of CLEVR-$2$ images, that does not contain
images with four spheres and a subset of CLEVR-$3$ without images of two cylinders, while at test time we evaluate the 
regression network on exactly such images.
The observed MSE values during testing are shown in Table~\ref{fig:interpolationclevr}.
These results show the potential of the model to transfer the learned count four from cylinders to spheres on CLEVR-$2$ and the learned count two from spheres and cubes to the cylinder class for CLEVR-$3$ images.
For extrapolation experiments, we train the model without samples from an object class with the maximum defined count in the dataset.
Since the maximum count in CLEVR-$2$ is six and for CLEVR-$3$ is three, we consider a subset of CLEVR-$2$ images without six spheres and CLEVR-$3$ images without three cylinders.
The observed MSE values (columns 3 and 4 in Table~\ref{fig:interpolationclevr}) are comparable with the considered baseline (column 2) for both interpolation and extrapolation of object count.
This further confirms that the network is not merely memorizing the count number. 
\subsection{Street View House Numbers (SVHN)}
In this section, we consider real world images from noisy training data on the street view house numbers (SVHN) dataset~\cite{svhn}.
The dataset includes house numbers cropped from street view images.
For our experiments, we considered the original images resized to $64 \times 64$ pixels and a total of $1500$ samples for each count combination. 
We restrict ourselves to SVHN images with at most two instances of each digit class (SVHN-$2$),
because images with three or more digits are too scarce for training. 
The count label is a vector of $10$ entries prescribing the multiplicity of each digit in the image.
The generated images are shown in Figure~\ref{fig:svhn_generatedstylegan}.

We observed an average count prediction accuracy of $93\%$, with an individual accuracy of $91\%$ for count one and $90\%$ for count two respectively 
(a more detailed analysis of the count prediction on SVHN is provided in the Appendix).
We frequently noticed incorrect labels in the original dataset which might affect the count label and prediction accuracy.
\subsection{CityCount}
\begin{figure}[h]
\centering
 \begin{minipage}{\columnwidth}
  \begin{tabular}{@{\hspace{0cm}}l@{\hspace{.1cm}}l@{\hspace{.0cm}}}
  \subcaptionbox{Cars. \label{subfig:histcars}}{\includegraphics[width=0.5\textwidth]{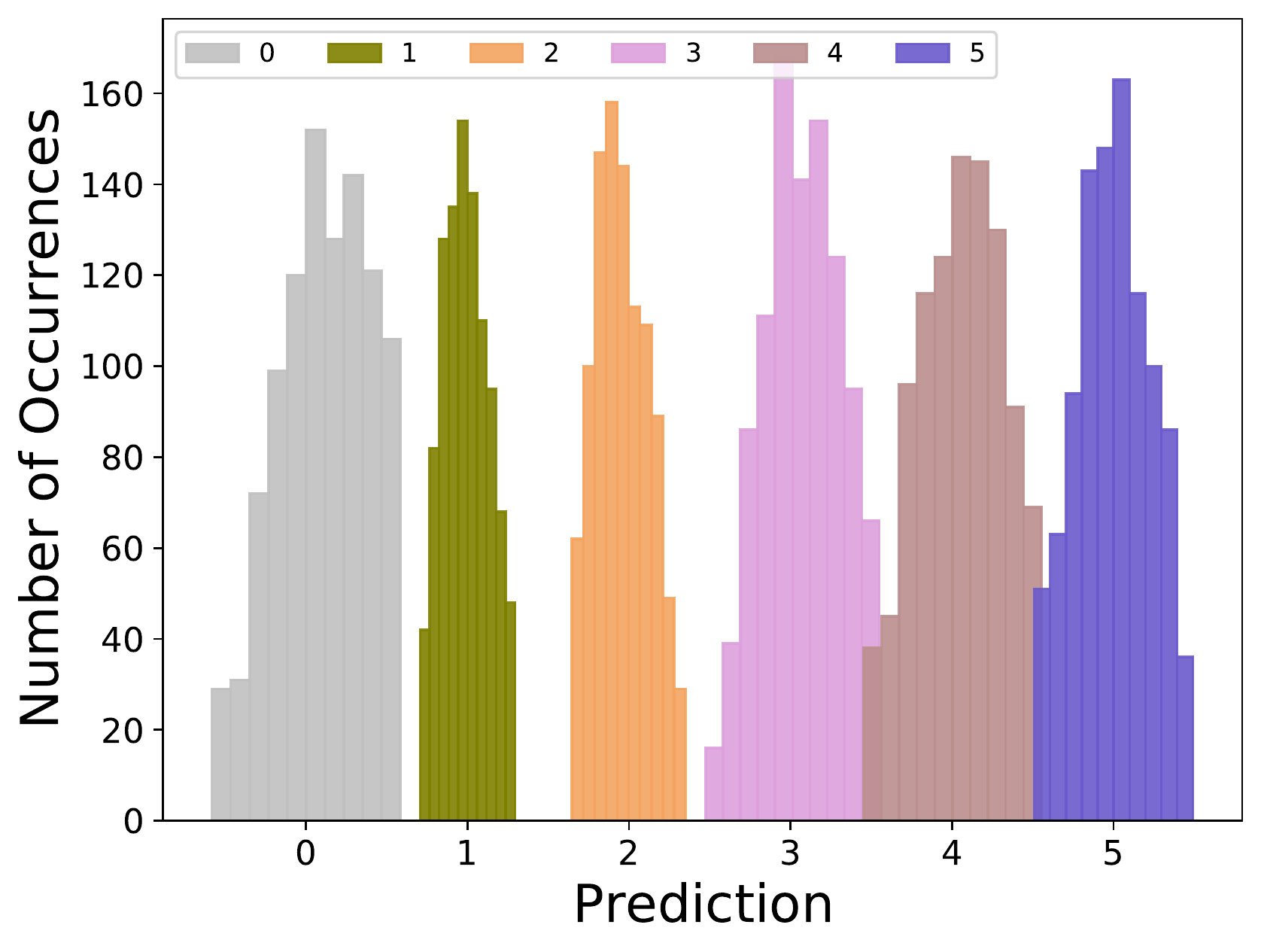}}&
  \subcaptionbox{Persons. \label{subfig:histperson}}{\includegraphics[width=0.5\textwidth]{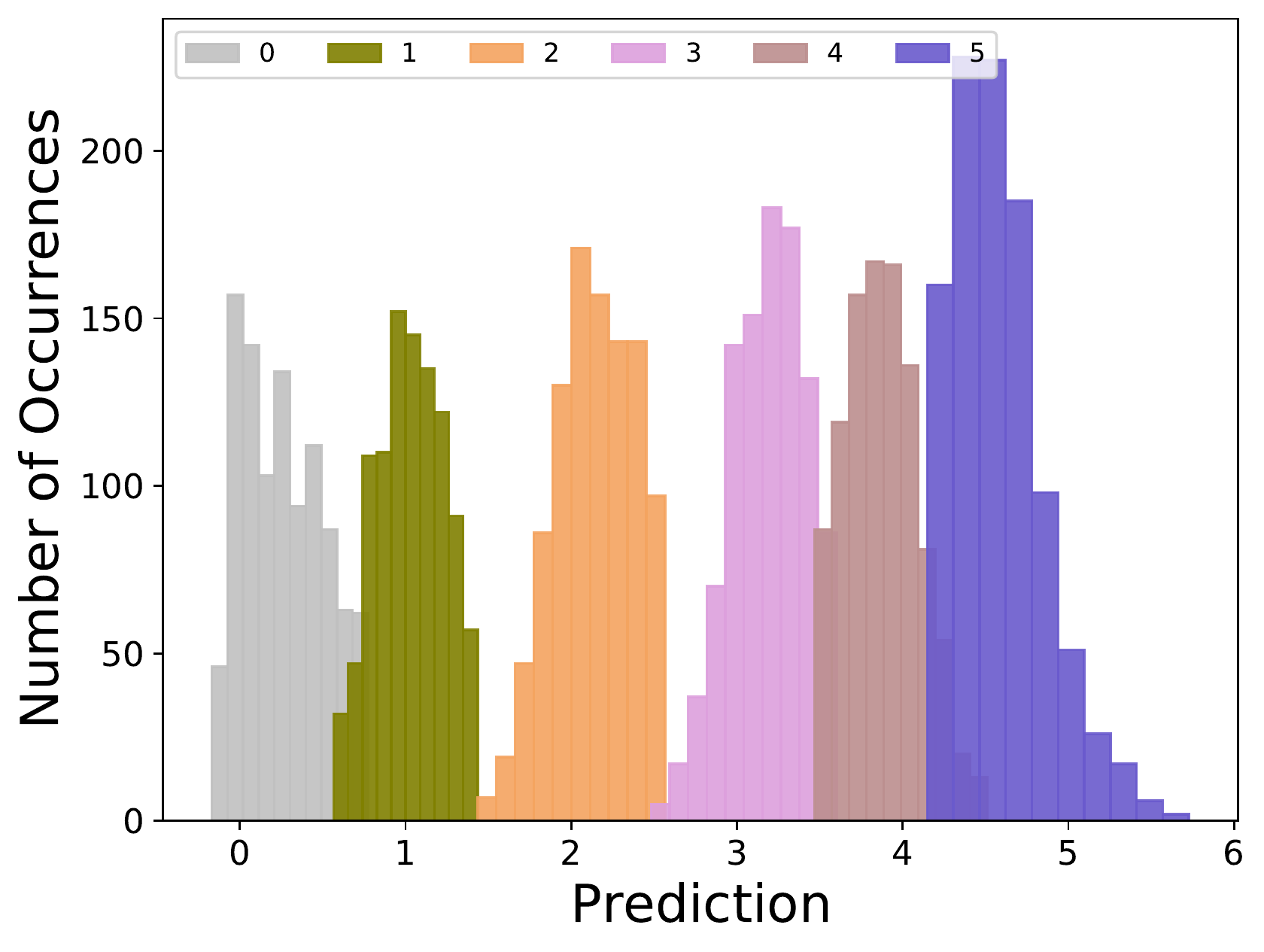}}
 \end{tabular}
 \end{minipage}\\
\caption{Count performance on CityCount generated images. The figure shows the predicted count values for car and person class of the generated samples from our model.} 
\label{fig:count_hist_Citycount}
\end{figure}
\noindent\textbf{Dataset} 
To evaluate our method on complex real world scenarios, we introduce a count based dataset derived from Cityscapes images, CityCount.
The images in CityCount are collected by cropping $256\times256$ size patches with defined number of \textit{cars} and \textit{persons} from Cityscapes. 
The dataset contains images with at most five instances from each of these classes and roughly $1000$ images per object class count combination.
To equip our dataset with additional count information, we determine the number of objects per class in each image from the $2$D bounding box information of cars and persons from the Cityscapes-3D~\cite{DBLP:journals/corr/abs-2006-07864} and the CityPerson dataset~\cite{Zhang2017CityPersonsAD}.
To allow for more diverse appearances of persons in the training set, classes including \textit{pedestrian}, \textit{sitting person} and \textit{rider} in the Cityscapes images are considered as positive samples when counting the number of persons in the images.
This further increases the complexity of the CityCount dataset in terms of spatial arrangement, since the network has to infer a reasonable placement of persons, like pedestrians on the sidewalk and riders on the road.
Since such additional spatial constraints are not explicitly specified, this makes our dataset more interesting and challenging for evaluating the proposed approach.
Most importantly, the bounding boxes, that were used to generate the training data, were not provided to the model during training.\\

\noindent\textbf{Evaluation} 
To account for the limited amount of training images, we used the adaptive discriminator augmentation technique~\cite{Karras2020ada} for training our model.
Samples of real and generated images with their respective count vector are shown in Figure~\ref{fig:citycount_generatedstylegan}.
Each count vector of size two represents the number of cars and persons.
For the ease of visualization, boxes are drawn around objects of interest.
The model generates images with diverse background and well defined person and car class placed spatially at reasonable locations.
As shown in the generated sample of 1 car and 2 persons combination in Figure~\ref{fig:citycount_generatedstylegan}, the person placed in the road can be seen along with a bike while the second person is standing on the sidewalk.
The model learns to distinguish between the pedestrian and the rider class even without an explicit definition of them in the training set.

We evaluate the predictive performance of the count sub-network for the car and person classes in Figure~\ref{subfig:histcars} and ~\ref{subfig:histperson} respectively.
Here, we compare the predicted count values on the generated samples with the true count provided to the generator network during test time.
Since in many samples of the training set persons are only partially visible and often out of focus or of low resolution, we observed a comparatively poor count performance for the person class.
For higher counts, $4$ or $5$, the relatively low performance is presumably due to the lower number of training samples and severe occlusions for the corresponding count.
\begin{table*}
\begin{center}
  \begin{tabular}{|c|cc|cc|cc|}
    \hline
     \multirow{3}{*}{Method} & \multicolumn{6}{c|}{Dataset} \\
     \cline{2-7}
     & \multicolumn{2}{c|}{CLEVR-$2$}  &\multicolumn{2}{c|}{CLEVR-$3$} &\multicolumn{2}{c|}{CityCount} \\
     \cline{2-7}
    &MSE($\downarrow$)
    &FID($\downarrow$) 
    &MSE($\downarrow$)
    &FID($\downarrow$) 
    &MSE($\downarrow$) 
    &FID($\downarrow$)
 \\
    \hline\hline
  w/o Count loss  &$1.377$  &$18.67$  &$1.401$  &$30.34$  &$1.129$ & $20.24$\\
    w/o Discriminator weight sharing &$0.024$    &$33.42$   &$0.113$    &$31.03$  &$0.441$& $15.78$\\
    w/o Label mapping       &$0.090$    &$11.01$   &$0.105$    &$11.32$  &$0.759$& $8.84$\\
    Residual generator   &$0.008$    &$8.28$   &\bf{0.011}    &$11.94$  &$0.446$& $11.72$\\
    Output skip generator   &$0.007$    &$8.62$   &$0.015$    &$8.98$  &$0.435$& $10.71$\\
    MC$^2$-StyleGAN2(ours)   &\bf{0.004}    &\bf{7.98}  &$0.022$    &\bf{8.94}  &\bf{0.356} &\bf{8.33} \\
   \hline
  \end{tabular}
\end{center}
\caption{Ablation study across datasets based on Mean Squared Error (MSE) and Fr\'echet Inception Distance (FID). The table shows the validity of the proposed architecture choices in our method. \label{tab_ablation}}  
\end{table*}
We perform an ablation study on synthetic dataset CLEVR and the real dataset CityCount to verify the importance of the additional count loss, generator design, weight sharing in the discriminator and the conditioning methods in the generator.

\vspace{0.3cm}
\noindent\textbf{Count loss}
We train our model without the count regression network and condition the generator and discriminator with the count label. The rest of our architecture is unchanged. The observed values (w/o count loss in Table 3 show that removing the count loss substantially degrades the performance both in terms of count prediction and image quality.

\vspace{0.3cm}
\noindent\textbf{Generator architecture}
We consider two different generator configurations introduced in StyleGAN2. One that uses output skip connections and a second one that uses residual connections. 
As shown in Table 3 (residual and output skip generator), our proposed dense like connections achieves overall good performance in terms of both count prediction and image quality.

\vspace{0.3cm}
\noindent\textbf{Weight sharing in the Discriminator} 
Further, we compute the evaluation metrics for our model without weight sharing between the count sub-network and the discriminator. 
The observed values in Table 3 (w/o discriminator weight sharing) show that the model failed to generate the object count correctly. This confirms the positive impact of weight sharing to regularize the count information and inform 
the discriminator.
\vspace{0.3cm}

\noindent\textbf{Count conditioning in Generator}
Lastly, we consider the setting where the count vector is not concatenated to every layer in the mapping network in the generator.
The results in Table 3 (w/o label mapping) show that the predictive performance is degraded in this setting.
This confirms the benefit of using a count vector based mapping network to propagate the multiple-class count effectively during training.  

\section{Comparison with other Methods}
\begin{table*}
\begin{center}
  \begin{tabular}{|c|cc|cc|cc|}
    \hline
     \multirow{3}{*}{Method} & \multicolumn{6}{c|}{Dataset} \\
     \cline{2-7}
     & \multicolumn{2}{c|}{CLEVR-$2$}  &\multicolumn{2}{c|}{CLEVR-$3$} &\multicolumn{2}{c|}{ SVHN-$2$ }  \\
     \cline{2-7}
    &MSE($\downarrow$)
    &FID($\downarrow$) 
    &MSE($\downarrow$)
    &FID($\downarrow$) 
    &MSE($\downarrow$) 
    &FID($\downarrow$)
 \\
    \hline\hline
   CGAN  &$6.924$  &$119.23$  &$2.919$  &$186.13$  &$0.301$ & $170.80$   \\
    ACGAN &$5.715$  &$99.88$  &$2.818$  &$132.23$  &$0.233$ & $150.56$  \\
    TACGAN       &$5.022$    &$92.04$   &$2.677$    &$120.11$  &$0.209$& $138.29$  \\
    CGAN(ourG)       &$1.259$    &$88.79$   &$1.106$    &$152.56$  &$0.141$& $90.34$  \\
    ACGAN(ourG)      &$1.164$    &$55.24$   &$1.176$    &$91.02$  &$0.139$& $70.28$  \\
    TACGAN(ourG)       &$1.129$    &$49.01$   &$1.091$    &$87.64$  &$0.135$& $65.77$  \\
    MC$^2$-SimpleGAN(ours)   &\underline{0.097}    &\underline{47.95}   &\underline{0.098}    &\underline{85.48} & \underline{0.047} & \underline{57.52}  \\
    \hline
    MC$^2$-StyleGAN2(ours)   &\bf{0.004}    &\bf{7.98}  &\bf{0.022}     &\bf{8.94}  &\bf{0.015} & \bf{10.90}  \\
    \hline
  \end{tabular}
\end{center}
\caption{Comparison with other methods across datasets based on Mean Squared Error (MSE) and Fr\'echet Inception Distance (FID). Underlined values denotes the proposed method performance on simple (MC$^2$-SimpleGAN) and bold values with complex architecture (MC$^2$-StyleGAN2).\label{tab_comparison}}  
\end{table*}
We compare the quantitative performance of other conditional GAN variants, CGAN~\cite{cgan}, ACGAN~\cite{acgan} and TACGAN~\cite{tacgan}, for multiple-class counting on CLEVR and SVHN images.
In order to have a fair comparison of our method with these conditional GAN variants, we use a less evolved network architecture in our proposed model. We call this simplified version of our approach, MC$^2$-SimpleGAN.
The MC$^2$-SimpleGAN generator gets as input a combination of randomly sampled noise and a multiple-class count vector.
The generator architecture is inspired by Densenet
architecture~\cite{densenet} and includes two dense blocks (where the output from each layer is connected in a feed forward fashion to its succeeding layers) followed by two fully connected layers.
The discriminator includes a convolutional based adversarial network and a count regression network with weight sharing.
For more architectural details please refer to the Appendix.

The initial results (row 1 to 3 in Table~\ref{tab_comparison}) indicate that the considered conditional GAN models did not perform well both in terms of image quality and FID.
We even observed mode collapse for CGAN.
Hence, we replaced the generator architecture of these models with the a Densenet based generator to improve the performance (rows 4 to 6 in Table~\ref{tab_comparison}).
Although we could greatly improve the initial performance of these models (which shows the positive impact of the proposed Densenet based generator), MC$^2$-SimpleGAN clearly outperforms other methods in the envisioned setting.
Further the quality of the generated images is improved with the proposed MC$^2$-StyleGAN2.
\section{Training Count Prediction Network using Synthetic Images}
\begin{table}
\centering
  \begin{tabular}{|c|c|c|c|}
    \hline
    \multirow{2}{*}{Training data}  & \multicolumn{3}{c|}{Accuracy($\uparrow$)}    \\ 
     \cline{2-4}
    &CLEVR
    &CityCount
    &CityCar
    
 \\
    \hline\hline
    Real only  & $0.81$  &$0.68$ &$0.77$   \\
    Real + Aug  &$0.81$  &\bf{0.71}  &$0.78$    \\
    Real + Syn(ours)   &\bf{0.86} &\bf{0.71} &\bf{0.80}       \\
    
    \hline
     \end{tabular}%
\caption{Average count accuracy across datasets for different training data setting. We used same number of augmented and synthetic images.\label{tab_count}
}  
\end{table}
We further demonstrate the usability of the images generated by  MC$^2$-StyleGAN2 for training a count prediction network.
In particular, we use a multiple-class extension of regression-based architecture similar to the discriminator of MC$^2$-SimpleGAN . 
The network aims to predict the number of objects per object class for the corresponding input images.
We design two experiments in this setting using CLEVR and CityCount images.
Since the quality of person instances in CityCount images is comparatively low, we also consider a subset of CityCount called CityCar, comprising solely of car class.
The average count accuracy of the model is considered as the evaluation metric.

In the first experiment, we evaluate whether 
the generated images can improve the count performance when combined along with real images during training. 
For baseline comparison, the count prediction network is initially trained with real images alone (first row in Table~\ref{tab_count}).
The network is then trained with a combination of real and augmented real images (second row in Table~\ref{tab_count}).
The observed count accuracy is then compared with the performance of the network when trained with real and the generated images (third row in Table~\ref{tab_count}).
For fair comparison we consider equal number of augmented and synthetic images.
As shown in Table~\ref{tab_count} for CLEVR and CityCar images the combination of real and synthetic images (Real+Syn) improved the baseline setting (Real only) and the combination of real and augmented images (Real+Aug). 
For CityCount, similar count performance is observed for both Real+Aug and Real+Syn.
\begin{table}
\centering
{\def\arraystretch{1.35}\tabcolsep=5pt
  \begin{tabular}{|c |c |c |c |@{}}
    \hline
    \multirow{2}{*}{Training data}     & \multicolumn{3}{c|}{Accuracy($\uparrow$)}\\ 
     \cline{2-4}
    &CLEVR
    &CityCount
    &CityCar
 \\
    \hline\hline
    Real only  & $0.81$  &\bf{0.68} &$0.75$   \\
\hline
    Syn(ours) only  & $0.40$  &$0.30$ &$0.39$   \\
 \hline
    25\% Real only   &{0.65} &{0.41} &{0.59}       \\
\hline
    25\% Real +   &\multirow{2}{*}{0.67} &\multirow{2}{*}{0.45} &\multirow{2}{*}{0.62}       \\
    75\%  Syn(ours) & & & \\
\hline
    50\% Real only  &{0.76} &{0.56} &{0.69}       \\
 \hline
    50\% Real +   &\multirow{2}{*}{0.81} &\multirow{2}{*}{0.60} &\multirow{2}{*}{0.75} \\
    50\%  Syn(ours)  & & & \\
\hline
    75\% Real only  &{0.77} &{0.65} &{0.74}       \\
 \hline
    75\% Real +    &\multirow{2}{*}{\bf{0.83}} &\multirow{2}{*}{\bf{0.68}} &\multirow{2}{*}{\bf{0.76}} \\
    25\%  Syn(ours)  & & & \\
    \hline
     \end{tabular}%
     }
\caption{Average count accuracy across datasets when count prediction network trained with real and images from MC$^2$-StyleGAN2 (Syn) at various proportions. \label{tab:countrealfakecomparison}
}  
\end{table}

In the second experiment, we investigate the potential of the generated images to replace the real images during training, without compromising the count accuracy performance.
We consider the setting where the network is trained with a combination of real and synthetic images at various ratios.
Initially, the network is trained with only real images and then with only synthetic images.
We gradually replace the real images with synthetic images at various proportions and evaluate the count performance for each setting as shown in Table~\ref{tab:countrealfakecomparison}. 
For the baseline comparison of each setting, we consider the count accuracy of the network when trained with the corresponding ratio of the real images only (x\% Real only in Table~\ref{tab:countrealfakecomparison}).
As seen in Table~\ref{tab:countrealfakecomparison}, 50\% of real images could be replaced by the generated images without compromising the overall count performance for both CLEVR and CityCar images.
The synthetic images could also improve the overall count performance of the network while replacing 25\% of real images for both CLEVR and CityCar images.
For CityCount images, 25\% of real images could be replaced by the generated images without compromising the overall count performance.
\section{Conclusion}
In this paper, we investigate the potential of GANs
to guide the image generation process based on the number of objects of different classes in the images. 
While the task of counting is in general very challenging for deep learning approaches, our proposed method can generate images based on the multiple-class count vector
in synthetic and real world datasets. 
Our experiments indicate that the numerosity of objects in the images provides strong information regarding their distinguishability during feature learning and hence allows control of the image generation process.
Our evaluation further shows that the model is able to interpolate and extrapolate to unseen counts for specific classes. 
Even without providing additional information such as the locations of objects in the image, the network infers a reasonable spatial layout and realization of the objects from the training data distribution solely using the count information.
{\small
\bibliographystyle{ieee_fullname}
\bibliography{references}

\begin{thebibliography}{10}\itemsep=-1pt

\bibitem{casualvqa}
V. Agarwal, Rakshith Shetty, and M. Fritz.
\newblock Towards causal vqa: Revealing and reducing spurious correlations by
  invariant and covariant semantic editing.
\newblock {\em 2020 IEEE/CVF Conference on Computer Vision and Pattern
  Recognition (CVPR)}, pages 9687--9695, 2020.

\bibitem{NIPS2018_7978}
Namrata Anand and Possu Huang.
\newblock Generative modeling for protein structures.
\newblock In S. Bengio, H. Wallach, H. Larochelle, K. Grauman, N. Cesa-Bianchi,
  and R. Garnett, editors, {\em Advances in Neural Information Processing
  Systems 31}, pages 7494--7505. Curran Associates, Inc., 2018.

\bibitem{vqa}
S. {Antol}, A. {Agrawal}, J. {Lu}, M. {Mitchell}, D. {Batra}, C.~L. {Zitnick},
  and D. {Parikh}.
\newblock Vqa: Visual question answering.
\newblock In {\em 2015 IEEE International Conference on Computer Vision
  (ICCV)}, pages 2425--2433, 2015.

\bibitem{text2video}
Yogesh Balaji, Martin Min, Bing Bai, Rama Chellappa, and Hans Graf.
\newblock Conditional gan with discriminative filter generation for
  text-to-video synthesis.
\newblock pages 1995--2001, 08 2019.

\bibitem{brock2018BigGAN}
Andrew Brock, Jeff Donahue, and Karen Simonyan.
\newblock Large scale gan training for high fidelity natural image synthesis.
\newblock In {\em International Conference on Learning Representations (ICLR)},
  2018.

\bibitem{segmentcount}
A.~B. {Chan}, {Zhang-Sheng John Liang}, and N. {Vasconcelos}.
\newblock Privacy preserving crowd monitoring: Counting people without people
  models or tracking.
\newblock In {\em 2008 IEEE Conference on Computer Vision and Pattern
  Recognition}, pages 1--7, 2008.

\bibitem{Chen2018EfficientAA}
Yuhua Chen, Feng Shi, Anthony~G. Christodoulou, Zhengwei Zhou, Yibin Xie, and
  Debiao Li.
\newblock Efficient and accurate mri super-resolution using a generative
  adversarial network and 3d multi-level densely connected network.
\newblock In {\em MICCAI}, 2018.

\bibitem{Cordts2016Cityscapes}
Marius Cordts, Mohamed Omran, Sebastian Ramos, Timo Rehfeld, Markus Enzweiler,
  Rodrigo Benenson, Uwe Franke, Stefan Roth, and Bernt Schiele.
\newblock The cityscapes dataset for semantic urban scene understanding.
\newblock In {\em Proc. of the IEEE Conference on Computer Vision and Pattern
  Recognition (CVPR)}, 2016.

\bibitem{numbersense}
S. Dehaene.
\newblock {\em The number sense: How the mind creates mathematics}.
\newblock Oxford University Press, 2011.

\bibitem{vaeunet}
Patrick Esser, Ekaterina Sutter, and Bj{\"o}rn Ommer.
\newblock A variational u-net for conditional appearance and shape generation.
\newblock {\em 2018 IEEE/CVF Conference on Computer Vision and Pattern
  Recognition}, pages 8857--8866, 2018.

\bibitem{densitycount}
L. {Fiaschi}, U. {Koethe}, R. {Nair}, and F.~A. {Hamprecht}.
\newblock Learning to count with regression forest and structured labels.
\newblock In {\em Proceedings of the 21st International Conference on Pattern
  Recognition (ICPR2012)}, pages 2685--2688, 2012.

\bibitem{cellcountdetection}
Giselle Flaccavento, Victor~S. Lempitsky, Iestyn Pope, Paul~R. Barber, Andrew
  Zisserman, J.~Alison Noble, and Boris Vojnovic.
\newblock Learning to count cells: Applications to lens-free imaging of large
  fields.
\newblock 2011.

\bibitem{DBLP:journals/corr/abs-2006-07864}
Nils G{\"{a}}hlert, Nicolas Jourdan, Marius Cordts, Uwe Franke, and Joachim
  Denzler.
\newblock Cityscapes 3d: Dataset and benchmark for 9 dof vehicle detection.
\newblock {\em CoRR}, abs/2006.07864, 2020.

\bibitem{tacgan}
Mingming Gong, Yanwu Xu, Chunyuan Li, Kun Zhang, and Kayhan Batmanghelich.
\newblock Twin auxiliary classifiers gan.
\newblock {\em Advances in neural information processing systems},
  32:1328--1337, 12 2019.

\bibitem{gan}
Ian~J. Goodfellow, Jean Pouget-Abadie, Mehdi Mirza, Bing Xu, David
  Warde-Farley, Sherjil Ozair, Aaron Courville, and Yoshua Bengio.
\newblock Generative adversarial nets.
\newblock In {\em Proceedings of the 27th International Conference on Neural
  Information Processing Systems - Volume 2}, NIPS’14, page 2672–2680,
  Cambridge, MA, USA, 2014. MIT Press.

\bibitem{gulrajani2017wgan}
Ishaan Gulrajani, Faruk Ahmed, Martin Arjovsky, Vincent Dumoulin, and Aaron~C
  Courville.
\newblock Improved training of wasserstein gans.
\newblock In {\em Advances in neural information processing systems (NeurIPS)},
  pages 5767--5777, 2017.

\bibitem{AttGan}
Z. {He}, W. {Zuo}, M. {Kan}, S. {Shan}, and X. {Chen}.
\newblock Attgan: Facial attribute editing by only changing what you want.
\newblock {\em IEEE Transactions on Image Processing}, 28(11):5464--5478, Nov
  2019.

\bibitem{spatialgan}
Tobias Hinz, Stefan Heinrich, and Stefan Wermter.
\newblock Generating multiple objects at spatially distinct locations.
\newblock In {\em International Conference on Learning Representations}, 2019.

\bibitem{densenet}
G. {Huang}, Z. {Liu}, L. {Van Der Maaten}, and K.~Q. {Weinberger}.
\newblock Densely connected convolutional networks.
\newblock In {\em 2017 IEEE Conference on Computer Vision and Pattern
  Recognition (CVPR)}, pages 2261--2269, 2017.

\bibitem{huang2017AdaIN}
Xun Huang and Serge Belongie.
\newblock Arbitrary style transfer in real-time with adaptive instance
  normalization.
\newblock In {\em Proceedings of the IEEE International Conference on Computer
  Vision (ICCV)}, 2017.

\bibitem{pix2pixgan}
Phillip Isola, Jun-Yan Zhu, Tinghui Zhou, and Alexei~A. Efros.
\newblock Image-to-image translation with conditional adversarial networks.
\newblock {\em 2017 IEEE Conference on Computer Vision and Pattern Recognition
  (CVPR)}, pages 5967--5976, 2016.

\bibitem{clevr}
Justin Johnson, Bharath Hariharan, Laurens van~der Maaten, Li Fei-Fei, C.
  Zitnick, and Ross Girshick.
\newblock Clevr: A diagnostic dataset for compositional language and elementary
  visual reasoning.
\newblock pages 1988--1997, 2017.

\bibitem{karras2017ProGAN}
Tero Karras, Timo Aila, Samuli Laine, and Jaakko Lehtinen.
\newblock Progressive growing of gans for improved quality, stability, and
  variation.
\newblock {\em arXiv preprint arXiv:1710.10196}, 2017.

\bibitem{Karras2020ada}
Tero Karras, Miika Aittala, Janne Hellsten, Samuli Laine, Jaakko Lehtinen, and
  Timo Aila.
\newblock Training generative adversarial networks with limited data.
\newblock In {\em Proc. NeurIPS}, 2020.

\bibitem{karras2019StyleGAN}
Tero Karras, Samuli Laine, and Timo Aila.
\newblock A style-based generator architecture for generative adversarial
  networks.
\newblock In {\em Proceedings of the IEEE Conference on Computer Vision and
  Pattern Recognition (CVPR)}, 2019.

\bibitem{karras2019StyleGAN2}
Tero Karras, Samuli Laine, Miika Aittala, Janne Hellsten, Jaakko Lehtinen, and
  Timo Aila.
\newblock Analyzing and improving the image quality of stylegan.
\newblock {\em arXiv preprint arXiv:1912.04958}, 2019.

\bibitem{qacount}
Ankit Kumar, Ozan Irsoy, Peter Ondruska, Mohit Iyyer, James Bradbury, Ishaan
  Gulrajani, Victor Zhong, Romain Paulus, and Richard Socher.
\newblock Ask me anything: Dynamic memory networks for natural language
  processing.
\newblock In {\em Proceedings of The 33rd International Conference on Machine
  Learning}, volume~48 of {\em Proceedings of Machine Learning Research}, pages
  1378--1387, New York, New York, USA, 2016. PMLR.

\bibitem{liu2019stgan}
Ming Liu, Yukang Ding, Min Xia, Xiao Liu, Errui Ding, Wangmeng Zuo, and Shilei
  Wen.
\newblock Stgan: A unified selective transfer network for arbitrary image
  attribute editing.
\newblock In {\em IEEE Conference on Computer Vision and Pattern Recognition
  (CVPR)}, 2019.

\bibitem{cgan}
Mehdi Mirza and Simon Osindero.
\newblock Conditional generative adversarial nets.
\newblock 2014.

\bibitem{miyato2018spectral_norm}
Takeru Miyato, Toshiki Kataoka, Masanori Koyama, and Yuichi Yoshida.
\newblock Spectral normalization for generative adversarial networks.
\newblock {\em arXiv preprint arXiv:1802.05957}, 2018.

\bibitem{projectioncgan}
Takeru Miyato and Masanori Koyama.
\newblock cgans with projection discriminator.
\newblock {\em ArXiv}, abs/1802.05637, 2018.

\bibitem{svhn}
Yuval Netzer, Tao Wang, Adam Coates, Alessandro Bissacco, Bo Wu, and Andrew Ng.
\newblock Reading digits in natural images with unsupervised feature learning.
\newblock {\em NIPS}, 2011.

\bibitem{acgan}
Augustus Odena, Christopher Olah, and Jonathon Shlens.
\newblock Conditional image synthesis with auxiliary classifier gans.
\newblock In {\em ICML}, 2017.

\bibitem{10.1115/1.4044229}
Sangeun Oh, Yongsu Jung, Seongsin Kim, Ikjin Lee, and Namwoo Kang.
\newblock {Deep Generative Design: Integration of Topology Optimization and
  Generative Models}.
\newblock {\em Journal of Mechanical Design}, 141(11), 09 2019.
\newblock 111405.

\bibitem{dcgan}
Alec Radford, Luke Metz, and Soumith Chintala.
\newblock Unsupervised representation learning with deep convolutional
  generative adversarial networks.
\newblock {\em CoRR}, abs/1511.06434, 2015.

\bibitem{radford2015dcgan}
Alec Radford, Luke Metz, and Soumith Chintala.
\newblock Unsupervised representation learning with deep convolutional
  generative adversarial networks.
\newblock {\em arXiv preprint arXiv:1511.06434}, 2015.

\bibitem{text2image}
Scott Reed, Zeynep Akata, Xinchen Yan, Lajanugen Logeswaran, Bernt Schiele, and
  Honglak Lee.
\newblock Generative adversarial text to image synthesis.
\newblock In {\em International Conference on Machine Learning (ICML)}, 2016.

\bibitem{countrnn2}
Mengye Ren and Richard~S. Zemel.
\newblock End-to-end instance segmentation with recurrent attention.
\newblock {\em 2017 IEEE Conference on Computer Vision and Pattern Recognition
  (CVPR)}, pages 293--301, 2017.

\bibitem{countrnn}
Bernardino Romera-Paredes and Philip Hilaire~Sean Torr.
\newblock Recurrent instance segmentation.
\newblock {\em ArXiv}, abs/1511.08250, 2016.

\bibitem{deeplearningcount}
S. {Seguí}, O. {Pujol}, and J. {Vitrià}.
\newblock Learning to count with deep object features.
\newblock In {\em 2015 IEEE Conference on Computer Vision and Pattern
  Recognition Workshops (CVPRW)}, pages 90--96, 2015.

\bibitem{detectioncount}
Oliver Sidla, Yuriy Lypetskyy, Norbert Brandle, and Stefan Seer.
\newblock Pedestrian detection and tracking for counting applications in
  crowded situations.
\newblock pages 70 -- 70, 12 2006.

\bibitem{nalu}
Andrew Trask, Felix Hill, Scott~E. Reed, Jack~W. Rae, Chris Dyer, and Phil
  Blunsom.
\newblock Neural arithmetic logic units.
\newblock In {\em NeurIPS}, 2018.

\bibitem{babi}
Jason Weston, Antoine Bordes, Sumit Chopra, and Tomas Mikolov.
\newblock Towards ai-complete question answering: A set of prerequisite toy
  tasks.
\newblock {\em CoRR}, abs/1502.05698, 2016.

\bibitem{children1990}
Karen Wynn.
\newblock Children's understanding of counting.
\newblock {\em Cognition}, 36(2):155--193, 1990.

\bibitem{resnest}
Hang Zhang, Chongruo Wu, Zhongyue Zhang, Yi Zhu, Zhi-Li Zhang, Haibin Lin, Yu e
  Sun, Tong He, Jonas Mueller, R. Manmatha, Mengnan Li, and Alexander~J. Smola.
\newblock Resnest: Split-attention networks.
\newblock {\em ArXiv}, abs/2004.08955, 2020.

\bibitem{Zhang2017CityPersonsAD}
Shanshan Zhang, Rodrigo Benenson, and B. Schiele.
\newblock Citypersons: A diverse dataset for pedestrian detection.
\newblock {\em 2017 IEEE Conference on Computer Vision and Pattern Recognition
  (CVPR)}, pages 4457--4465, 2017.

\bibitem{cyclegan}
Jun-Yan Zhu, Taesung Park, Phillip Isola, and Alexei Efros.
\newblock Unpaired image-to-image translation using cycle-consistent
  adversarial networks.
\newblock pages 2242--2251, 10 2017.

\end{thebibliography}
}
\clearpage
\appendix
\counterwithin{figure}{section}\counterwithin{figure}{section}
\counterwithin{table}{section}\counterwithin{table}{section}
\renewcommand\thefigure{\thesection.\arabic{figure}}  
\noindent{\Large\textbf{Appendix}} \\\\
In this document, we provide additional details and results to the main paper. The document is structured as follows:
\begin{enumerate}[label=(\Alph*)]    
  \item Additional quantitative analysis
  \begin{itemize}
    \item CLEVR (same colour)
    \item Extended comparative analysis on projection based discriminator GAN
    \end{itemize}
\item Extended analysis across dataset
\begin{itemize}
    \item Multi-MNIST
    \item Extended qualitative analysis
    \item  Extended quantitative analysis
  \end{itemize}
\item Network architecture and implementation details
\begin{itemize}
    \item MC$^2$-SimpleGAN 
    \item MC$^2$-StyleGAN2 
    \item Count prediction network
    \end{itemize}
\end{enumerate}
\section{Additional quantitative analysis}
\subsection{CLEVR}
The quantitative evaluation for CLEVR simple images where similar shapes are constrained to same color (red cylinders, green spheres and blue cubes) are shown in Table~\ref{tab:CLEVRsimple}.
\begin{table}[h]
  \label{tab_ablation}
  \centering 
  \vspace{\baselineskip}
  \begin{tabular}{|c|c|c|}
    \hline
    Dataset     & MSE($\downarrow$) & FID($\downarrow$) \\
    \hline\hline
     CLEVR-$2$  &$0.005$ &$6.524$\\ 
    CLEVR-$3$    &$0.010$  &$8.023$ \\
    \hline
  \end{tabular}
  \caption{Mean Squared Error (MSE) and Fr\'echet Inception Distance (FID) - CLEVR simple dataset.}\label{tab:CLEVRsimple}
\end{table}
\subsection{Extended comparative analysis}
In this section, we extend our baseline evaluations of our approach by considering the performance of one of the state of the art conditional GAN based on projection based discriminator, SNGAN.
Technically, state of the art conditional GANs such as SNGAN and BigGAN considers conditioning as a classification problem rather than regression.
As mentioned in the main paper, counting should be considered as a regression problem than a classification one to transfer the count property from one class to another.
However to demonstrate this experimentally we adapted SNGAN to address multiple-class multiple-label classification.
The observed FID and MSE values are given in Table~\ref{tab:sngan}.
SNGAN performs well in terms of FID (when compared to MC$^2$-SimpleGAN) but is significantly worse in terms of MSE than the proposed approach.
Moreover in SNGAN, the transfer of count property from one class to another is not possible by its definition.
\begin{table}[h]
  \label{tab_ablation}
  \centering 
  \vspace{\baselineskip}
  \begin{tabular}{|c|c|c|}
    \hline
    Dataset     & \multicolumn{2}{c|}{SNGAN} \\
    \cline{2-3}
& MSE($\downarrow$) & FID($\downarrow$)
 \\
    \hline\hline
     CLEVR-$2$  &$1.121$ &$29.34$\\ 
    CLEVR-$3$    &$1.160$  &$43.68$ \\
    SVHN-$2$ &   $0.135$  &$47.34$ \\
    \hline
  \end{tabular}
  \caption{Mean Squared Error (MSE) and Fr\'echet Inception Distance (FID) - SNGAN.}\label{tab:sngan}
\end{table}
\section{Extended analysis across dataset}
\subsection{Multi MNIST}
\begin{figure*}
 \scriptsize
 \selectfont
 \centering
  \subcaptionbox{Generated MNIST-$2$ images. \label{subfig:mnist2_generated}}{
    \begin{tabular}{@{\hspace{.0cm}}c@{\hspace{.03cm}}c@{\hspace{.03cm}}c@{\hspace{.03cm}}c@{\hspace{.03cm}}c@{\hspace{.03cm}}c@{\hspace{.03cm}}c@{\hspace{.03cm}}c@{\hspace{.03cm}}c@{\hspace{.03cm}}c@{\hspace{.0cm}}
    }
 
\includegraphics[width=0.07\textwidth]{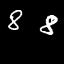} & 
 \includegraphics[width=0.07\textwidth]{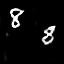} & 
 \includegraphics[width=0.07\textwidth]{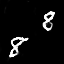} & 
 \includegraphics[width=0.07\textwidth]{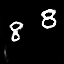} & 
 \includegraphics[width=0.07\textwidth]{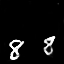} & 
 \includegraphics[width=0.07\textwidth]{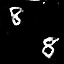} & 
 \includegraphics[width=0.07\textwidth]{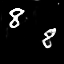} & 
 \includegraphics[width=0.07\textwidth]{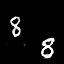} & 
 \includegraphics[width=0.07\textwidth]{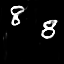} & 
 \includegraphics[width=0.07\textwidth]{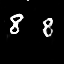} 
 \\
 \multicolumn{10}{@{\hspace{.01cm}}c@{\hspace{.01cm}}}{{[}0 0 0 0 0 0 0 0 2 0]} \\

\includegraphics[width=0.07\textwidth]{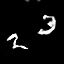} & 
 \includegraphics[width=0.07\textwidth]{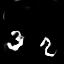} & 
 \includegraphics[width=0.07\textwidth]{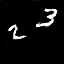} & 
 \includegraphics[width=0.07\textwidth]{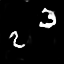} & 
 \includegraphics[width=0.07\textwidth]{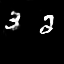} & 
 \includegraphics[width=0.07\textwidth]{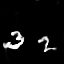} & 
 \includegraphics[width=0.07\textwidth]{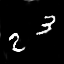} & 
 \includegraphics[width=0.07\textwidth]{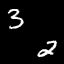} & 
 \includegraphics[width=0.07\textwidth]{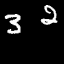} & 
 \includegraphics[width=0.07\textwidth]{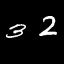} 
 \\
  \multicolumn{10}{@{\hspace{.01cm}}c@{\hspace{.01cm}}}{{[}0 0 1 1 0 0 0 0 0 0]}  \\

\includegraphics[width=0.07\textwidth]{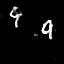} & 
 \includegraphics[width=0.07\textwidth]{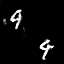} & 
 \includegraphics[width=0.07\textwidth]{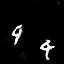} & 
 \includegraphics[width=0.07\textwidth]{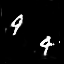} & 
 \includegraphics[width=0.07\textwidth]{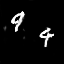} & 
 \includegraphics[width=0.07\textwidth]{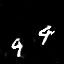} & 
 \includegraphics[width=0.07\textwidth]{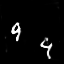} & 
 \includegraphics[width=0.07\textwidth]{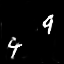} & 
 \includegraphics[width=0.07\textwidth]{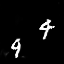} & 
 \includegraphics[width=0.07\textwidth]{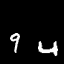} 
 \\
 \multicolumn{10}{@{\hspace{.01cm}}c@{\hspace{.01cm}}}{{[}0 0 0 0 1 0 0 0 0 1]}  \\

\includegraphics[width=0.07\textwidth]{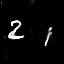} & 
 \includegraphics[width=0.07\textwidth]{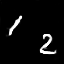} & 
 \includegraphics[width=0.07\textwidth]{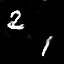} & 
 \includegraphics[width=0.07\textwidth]{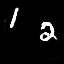} & 
 \includegraphics[width=0.07\textwidth]{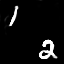} & 
 \includegraphics[width=0.07\textwidth]{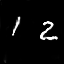} & 
 \includegraphics[width=0.07\textwidth]{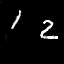} & 
 \includegraphics[width=0.07\textwidth]{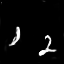} & 
 \includegraphics[width=0.07\textwidth]{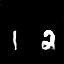} & 
 \includegraphics[width=0.07\textwidth]{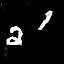} 
 \\
 \multicolumn{10}{@{\hspace{.01cm}}c@{\hspace{.01cm}}}{{[}0 1 1 0 0 0 0 0 0 0]}  \\
\includegraphics[width=0.07\textwidth]{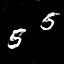} & 
 \includegraphics[width=0.07\textwidth]{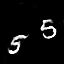} & 
 \includegraphics[width=0.07\textwidth]{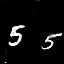} & 
 \includegraphics[width=0.07\textwidth]{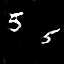} & 
 \includegraphics[width=0.07\textwidth]{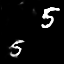} & 
 \includegraphics[width=0.07\textwidth]{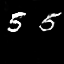} & 
 \includegraphics[width=0.07\textwidth]{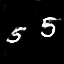} & 
 \includegraphics[width=0.07\textwidth]{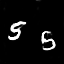} & 
 \includegraphics[width=0.07\textwidth]{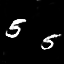} & 
 \includegraphics[width=0.07\textwidth]{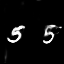} 
 \\
\multicolumn{10}{@{\hspace{.01cm}}c@{\hspace{.01cm}}}{{[}0 0 0 0 0 2 0 0 0 0]}  \\

\end{tabular}}
 \subcaptionbox{Generated MNIST-$3$ images. \label{subfig:mnist3_generated}}{
    \begin{tabular}{@{\hspace{.0cm}}c@{\hspace{.03cm}}c@{\hspace{.03cm}}c@{\hspace{.03cm}}c@{\hspace{.03cm}}c@{\hspace{.03cm}}c@{\hspace{.03cm}}c@{\hspace{.03cm}}c@{\hspace{.03cm}}c@{\hspace{.03cm}}c@{\hspace{.0cm}}
    }
   
\includegraphics[width=0.07\textwidth]{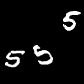} & 
 \includegraphics[width=0.07\textwidth]{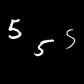} & 
 \includegraphics[width=0.07\textwidth]{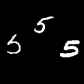} & 
 \includegraphics[width=0.07\textwidth]{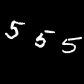} & 
 \includegraphics[width=0.07\textwidth]{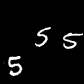} & 
 \includegraphics[width=0.07\textwidth]{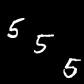} & 
 \includegraphics[width=0.07\textwidth]{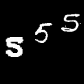} & 
 \includegraphics[width=0.07\textwidth]{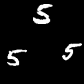} & 
 \includegraphics[width=0.07\textwidth]{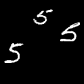} & 
 \includegraphics[width=0.07\textwidth]{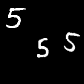} 
 \\
   \multicolumn{10}{@{\hspace{.01cm}}c@{\hspace{.01cm}}}{{[}0 0 0 0 0 3 0 0 0 0]}  \\

\includegraphics[width=0.07\textwidth]{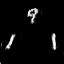} & 
 \includegraphics[width=0.07\textwidth]{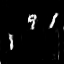} & 
 \includegraphics[width=0.07\textwidth]{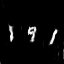} & 
 \includegraphics[width=0.07\textwidth]{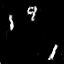} & 
 \includegraphics[width=0.07\textwidth]{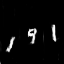} & 
 \includegraphics[width=0.07\textwidth]{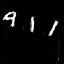} & 
 \includegraphics[width=0.07\textwidth]{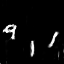} & 
 \includegraphics[width=0.07\textwidth]{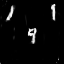} & 
 \includegraphics[width=0.07\textwidth]{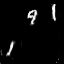} & 
 \includegraphics[width=0.07\textwidth]{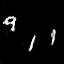} 
 \\
  \multicolumn{10}{@{\hspace{.01cm}}c@{\hspace{.01cm}}}{{[}0 2 0 0 0 0 0 0 0 1]}  \\

  \includegraphics[width=0.07\textwidth]{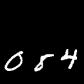} & 
\includegraphics[width=0.07\textwidth]{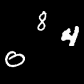} & 
 \includegraphics[width=0.07\textwidth]{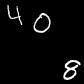} & 
 \includegraphics[width=0.07\textwidth]{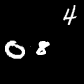} & 
  \includegraphics[width=0.07\textwidth]{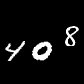} & 
   \includegraphics[width=0.07\textwidth]{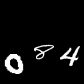} & 
 \includegraphics[width=0.07\textwidth]{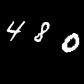} & 
 \includegraphics[width=0.07\textwidth]{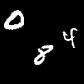} & 
 \includegraphics[width=0.07\textwidth]{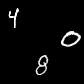} & 

 \includegraphics[width=0.07\textwidth]{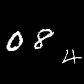} 
 \\
  \multicolumn{10}{@{\hspace{.01cm}}c@{\hspace{.01cm}}}{{[}1 0 0 0 1 0 0 0 1 0]}  \\

\includegraphics[width=0.07\textwidth]{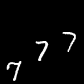} & 
 \includegraphics[width=0.07\textwidth]{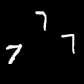} & 
 \includegraphics[width=0.07\textwidth]{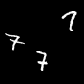} & 
 \includegraphics[width=0.07\textwidth]{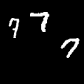} & 
 \includegraphics[width=0.07\textwidth]{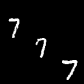} & 
 \includegraphics[width=0.07\textwidth]{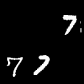} & 
 \includegraphics[width=0.07\textwidth]{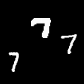} & 
 \includegraphics[width=0.07\textwidth]{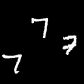} & 
 \includegraphics[width=0.07\textwidth]{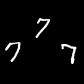} & 
 \includegraphics[width=0.07\textwidth]{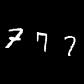} 
 \\
  \multicolumn{10}{@{\hspace{.01cm}}c@{\hspace{.01cm}}}{{[}0 0 0 0 0 0 0 3 0 0]} \\

\includegraphics[width=0.07\textwidth]{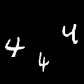} & 
 \includegraphics[width=0.07\textwidth]{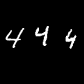} & 
 \includegraphics[width=0.07\textwidth]{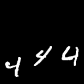} & 
 \includegraphics[width=0.07\textwidth]{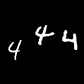} & 
 \includegraphics[width=0.07\textwidth]{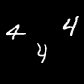} & 
 \includegraphics[width=0.07\textwidth]{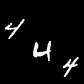} & 
 \includegraphics[width=0.07\textwidth]{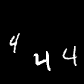} & 
 \includegraphics[width=0.07\textwidth]{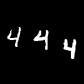} & 
 \includegraphics[width=0.07\textwidth]{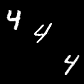} & 
 \includegraphics[width=0.07\textwidth]{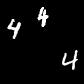} 
 \\
  \multicolumn{10}{@{\hspace{.01cm}}c@{\hspace{.01cm}}}{{[}0 0 0 0 3 0 0 0 0 0]}  \\
 
 \end{tabular}}   
 \subcaptionbox{Generated MNIST images for unseen count combination. \label{subfig:mnist_unseen}}{
    \begin{tabular}{@{\hspace{.0cm}}c@{\hspace{.03cm}}c@{\hspace{.03cm}}c@{\hspace{.03cm}}c@{\hspace{.03cm}}c@{\hspace{.03cm}}c@{\hspace{.03cm}}c@{\hspace{.03cm}}c@{\hspace{.03cm}}c@{\hspace{.0cm}}
    }
    \includegraphics[width=0.075\textwidth]{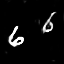} & 
 \includegraphics[width=0.075\textwidth]{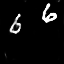} &
 \includegraphics[width=0.075\textwidth]{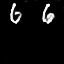} & 
 \includegraphics[width=0.075\textwidth]{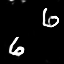} & 
 \includegraphics[width=0.075\textwidth]{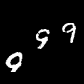} &
 \includegraphics[width=0.075\textwidth]{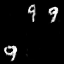} &
 \includegraphics[width=0.075\textwidth]{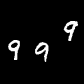} & 
 \includegraphics[width=0.075\textwidth]{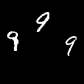} &
   \includegraphics[width=0.075\textwidth]{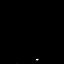} 
 \\
 \multicolumn{4}{@{\hspace{.01cm}}c@{\hspace{.01cm}}}{{[}0 0 0 0 0 0 2 0 0 0]} & \multicolumn{4}{@{\hspace{.01cm}}c@{\hspace{.01cm}}}{{[}0 0 0 0 0 0 0 0 0 3]} 
 & {all zeros} \\
    \end{tabular}}
 \caption{Generated Multi-MNIST images
 for different ten-dimensional count vectors - MC$^2$-SimpleGAN.}\label{fig:mnistimages}
 \end{figure*}
We consider Multi-MNIST dataset to evaluate the general capability of our approach to generate images based on learned count information using MC$^2$-SimpleGAN.
We evaluate on two variants of the Multi-MNIST dataset, one with two digits per image (MNIST-$2$) and the other with three digits per image (MNIST-$3$).
The datasets were generated by uniformly sampling digits from the MNIST dataset and placing them in non-overlapping positions on black background.
We used $1000$ images for each digit combination during training.
The count information is provided to the model as a vector with ten entries comprising the desired number of instances of each digit in the image.
Figure~\ref{fig:mnistimages} shows the generated samples by our model for different count combinations. 
The results show that the proposed model is able to produce images based on the given digit count without any supervision. \\

We observed an average count accuracy of $96\%$ for MNIST-$2$ images and $93\%$ for MNIST-$3$, where the accuracy slightly decreases for higher counts.
\\\\
\noindent\textbf{Interpolation and Extrapolation} To check for interpolation and extrapolation ability of the model, we trained the network with images containing only certain combinations of input count and during testing we input an unseen count combination of digits (see Figure~\ref{subfig:mnist_unseen}).
For the MNIST-$2$ dataset, the model sees the count value of only $1$ for digit $6$ during training.
Similarly for the MNIST-$3$ dataset, the count $3$ for digit $9$ is unseen during training.
The model is able to transfer the concept of count from one digit to another and predict the count for these unseen combinations. 
For count value $0$ for all digits (also unseen during training), the model is able to generate images without any digits.
\subsection{Extended quantitative analysis across datasets}
\begin{figure*}
 \subcaptionbox{CLEVR-$2$ \label{figHISTCLEVR2}}
  {\begin{tabular}{c@{\hspace{.45cm}}c@{\hspace{.45cm}}c@{\hspace{.45cm}}}
   \includegraphics[width=0.17\textwidth]{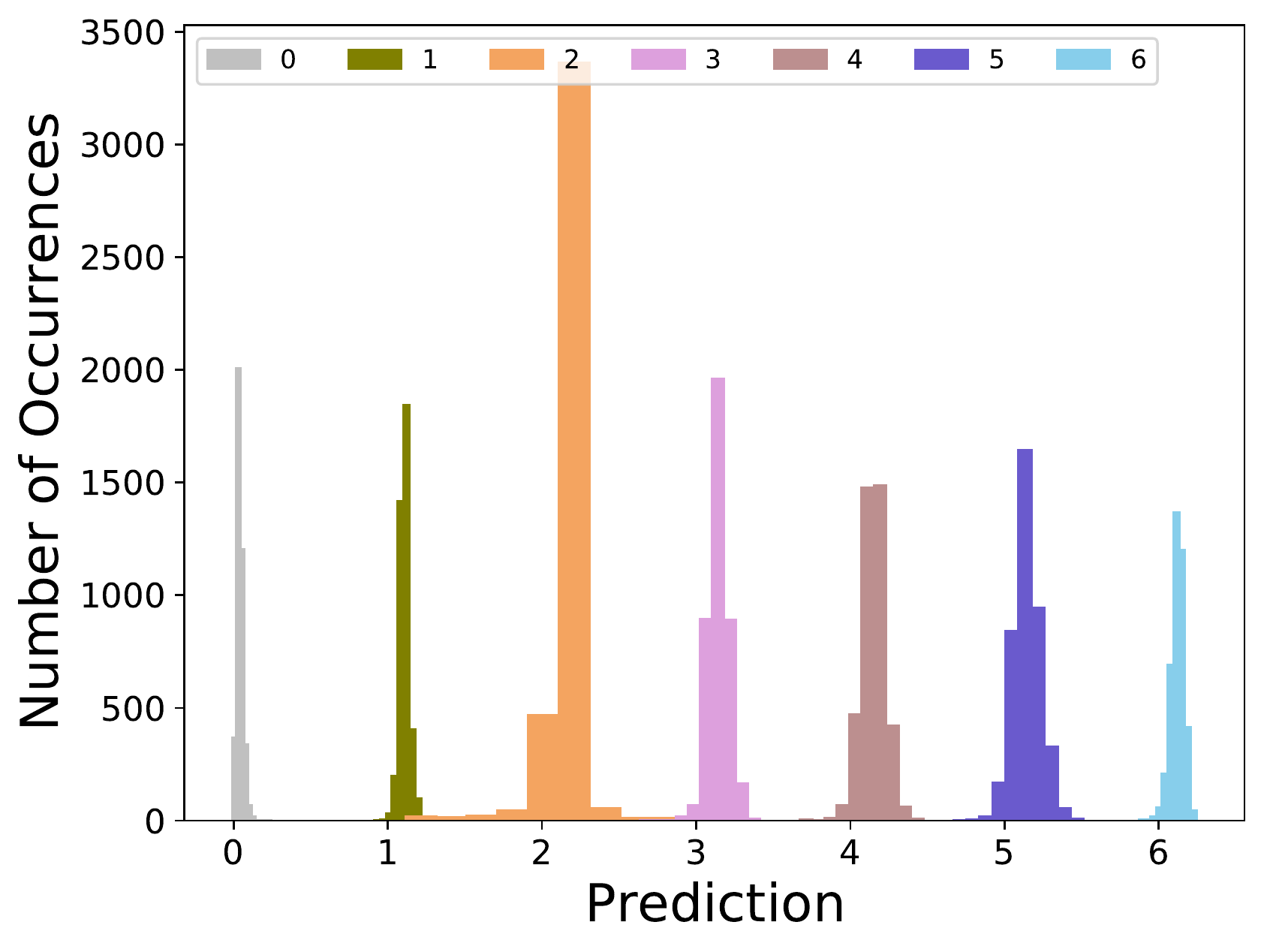}&
  \includegraphics[width=0.17\textwidth]{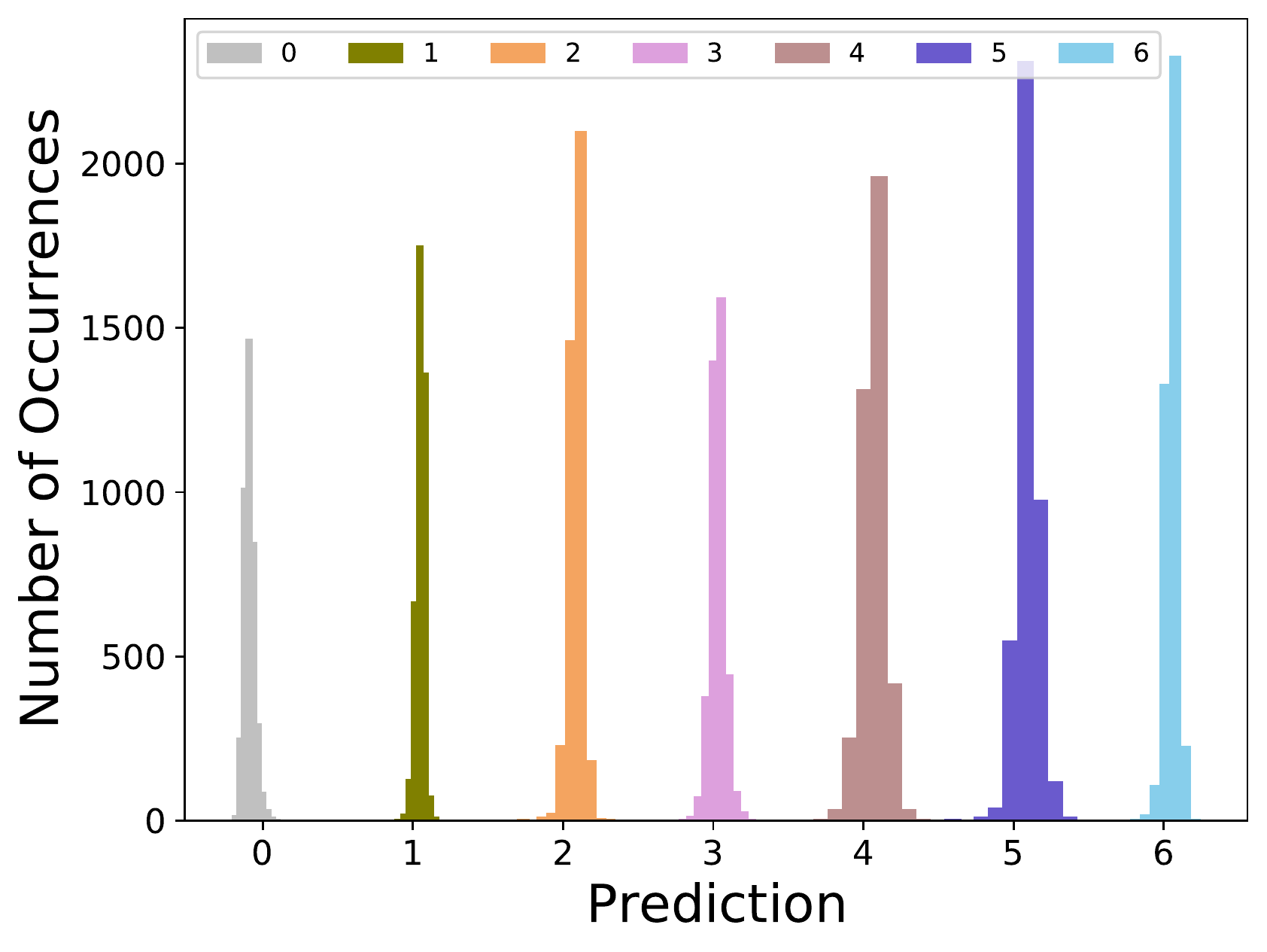} \\
  Cylinder &Sphere
 \end{tabular}
 }
 \subcaptionbox{CLEVR-$3$ \label{fig:HISTCLEVR3}}
 {\begin{tabular}{c@{\hspace{.45cm}}c@{\hspace{.45cm}}c@{\hspace{.45cm}}c@{\hspace{.0cm}}}

  \includegraphics[width=0.17\textwidth]{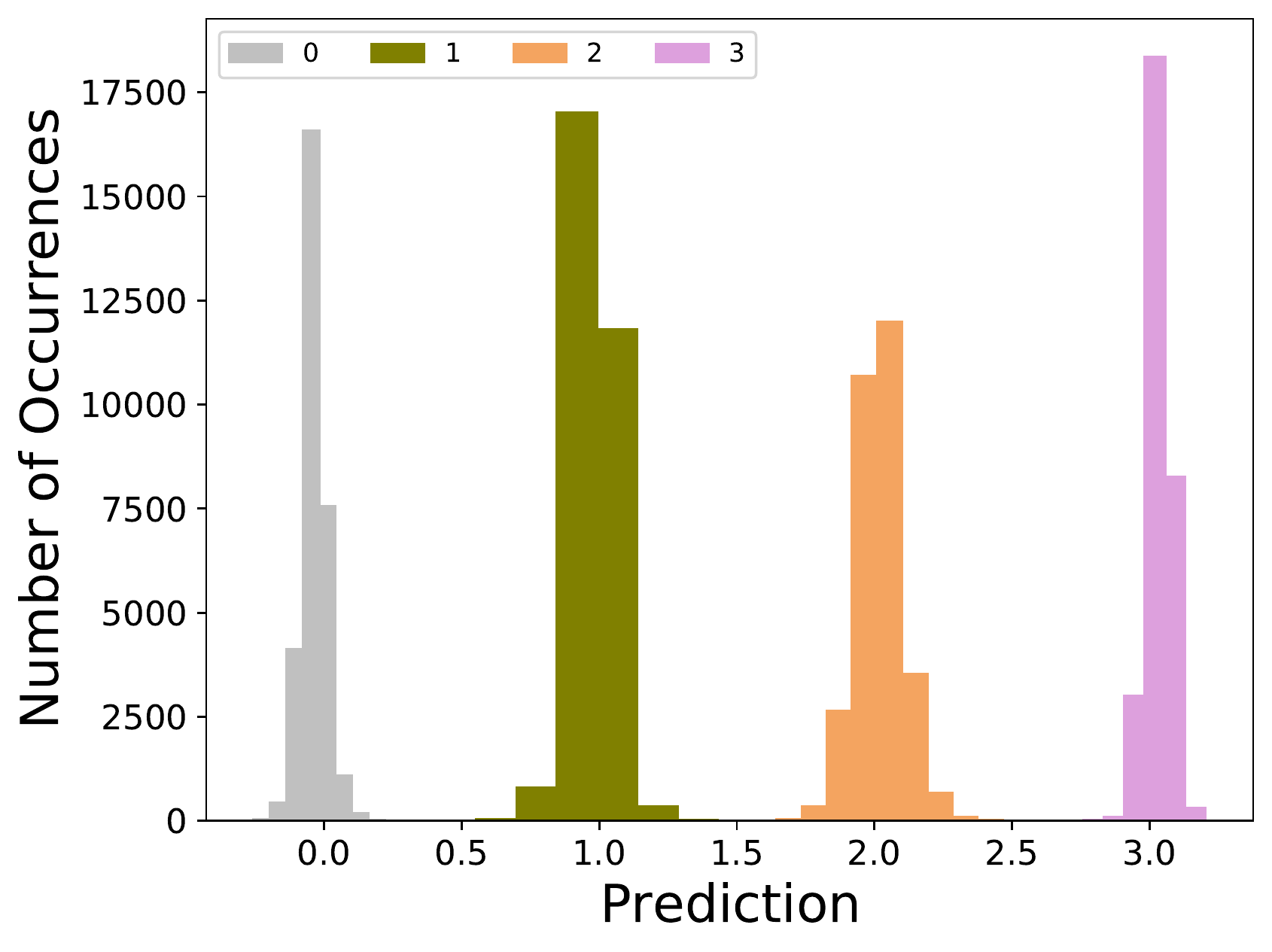}&
    \includegraphics[width=0.17\textwidth]{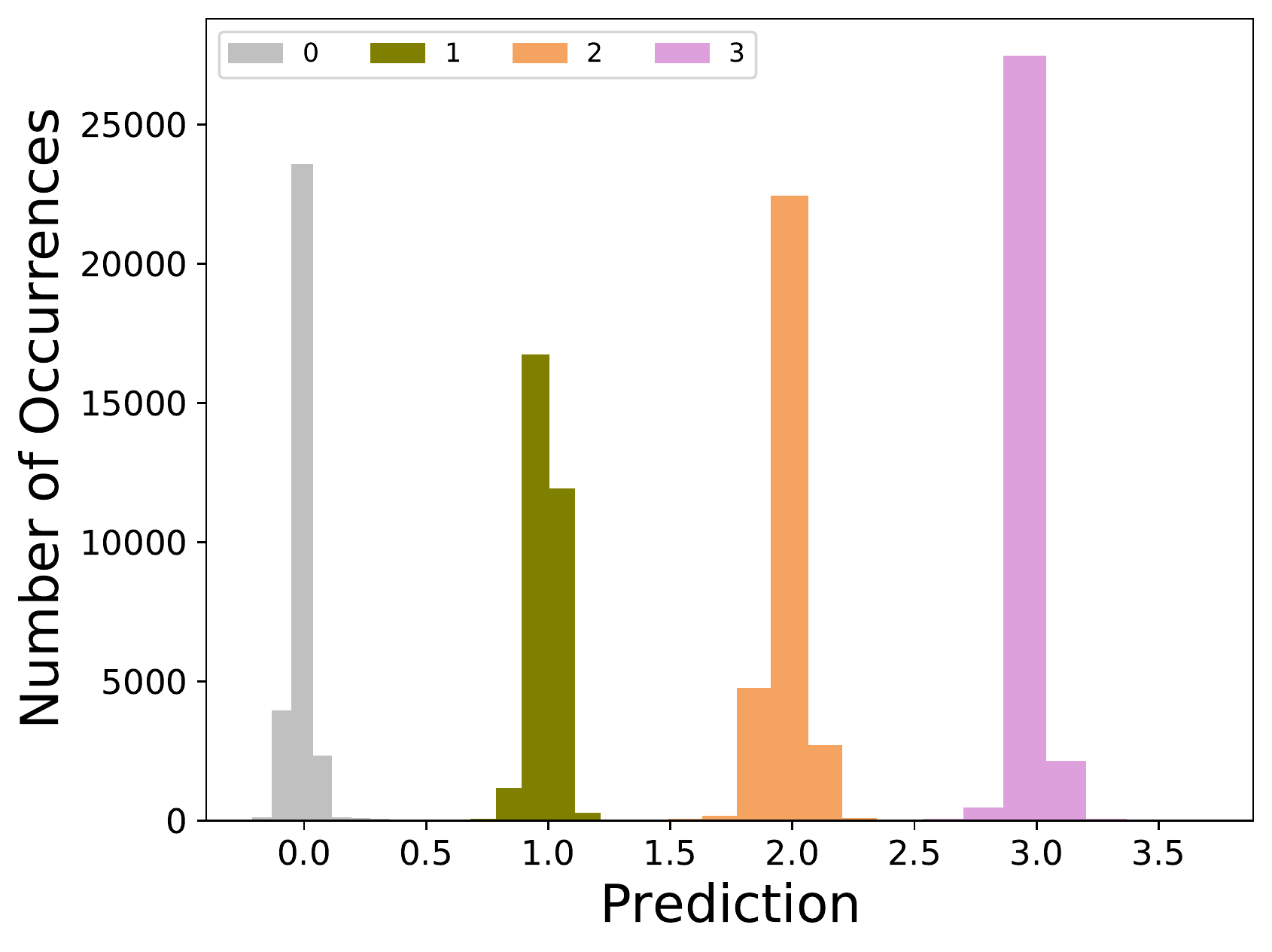} &
    \includegraphics[width=0.17\textwidth]{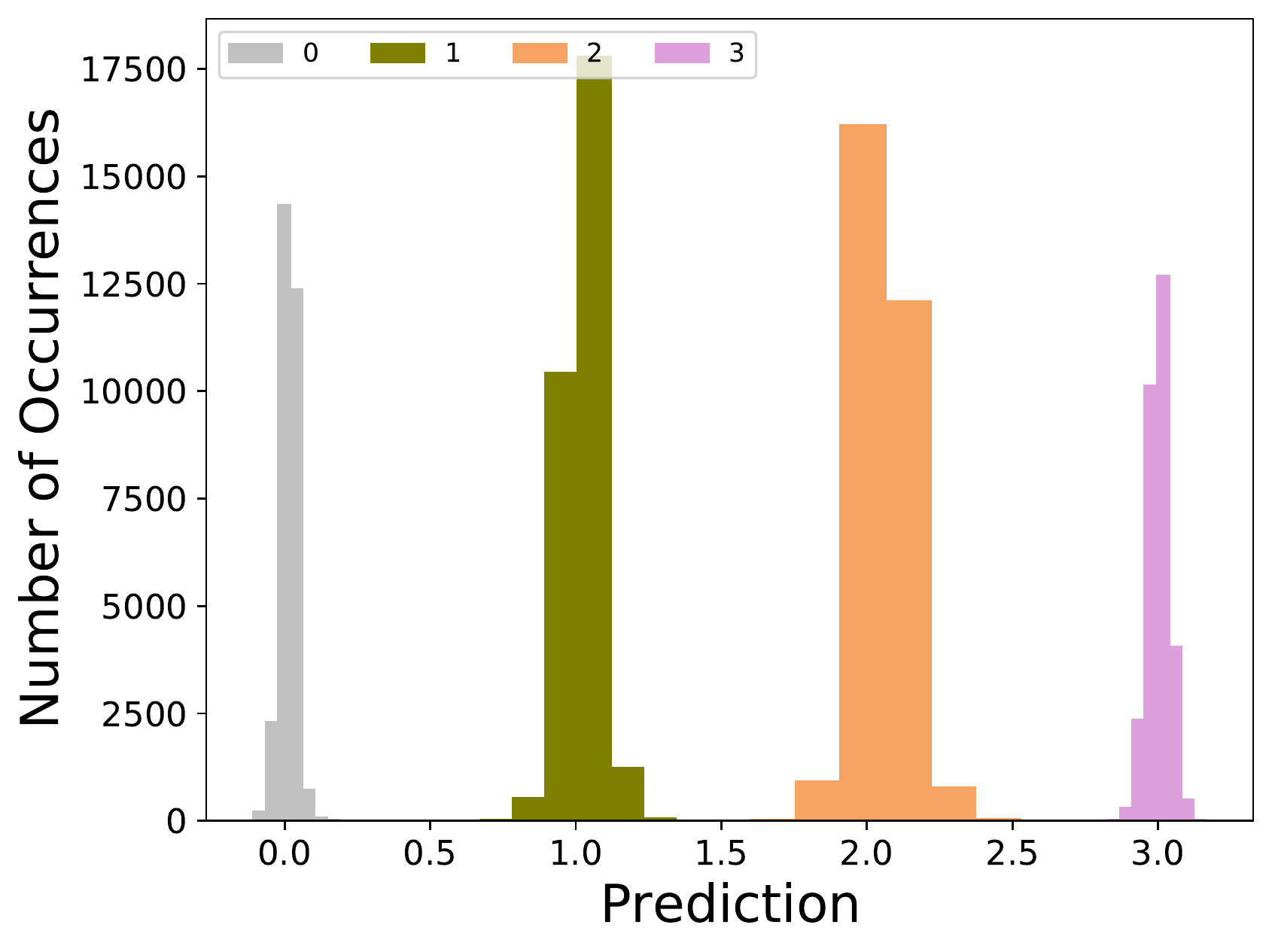}\\Cylinder &Sphere &Cube  
 \end{tabular}}
 \subcaptionbox{SVHN \label{fig:HISTSVHN}}
  {\begin{tabular}{c@{\hspace{.45cm}}c@{\hspace{.45cm}}c@{\hspace{.45cm}}c@{\hspace{.45cm}}c@{\hspace{.45cm}}c@{\hspace{.0cm}}}
   \includegraphics[width=0.17\textwidth]{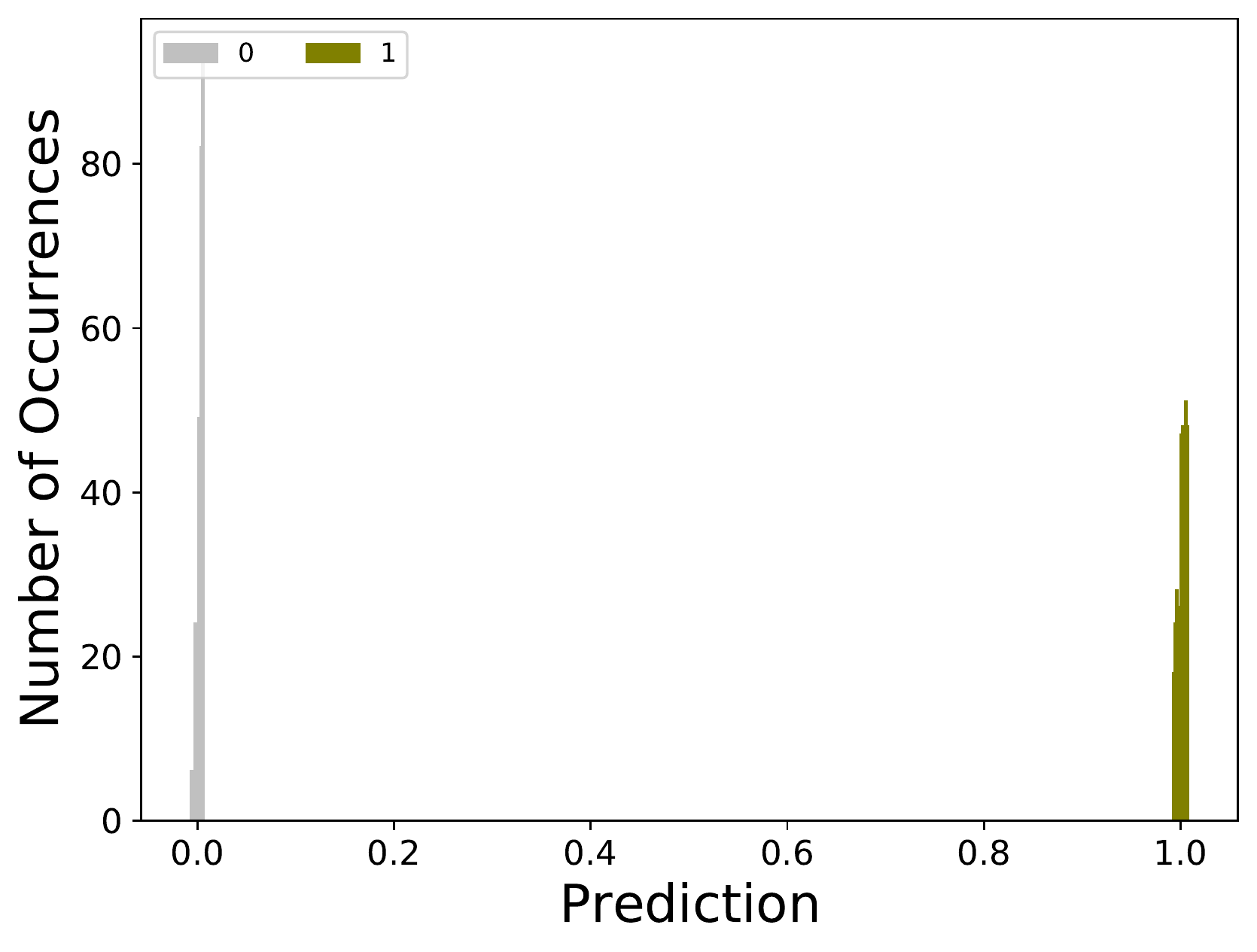}&
   \includegraphics[width=0.17\textwidth]{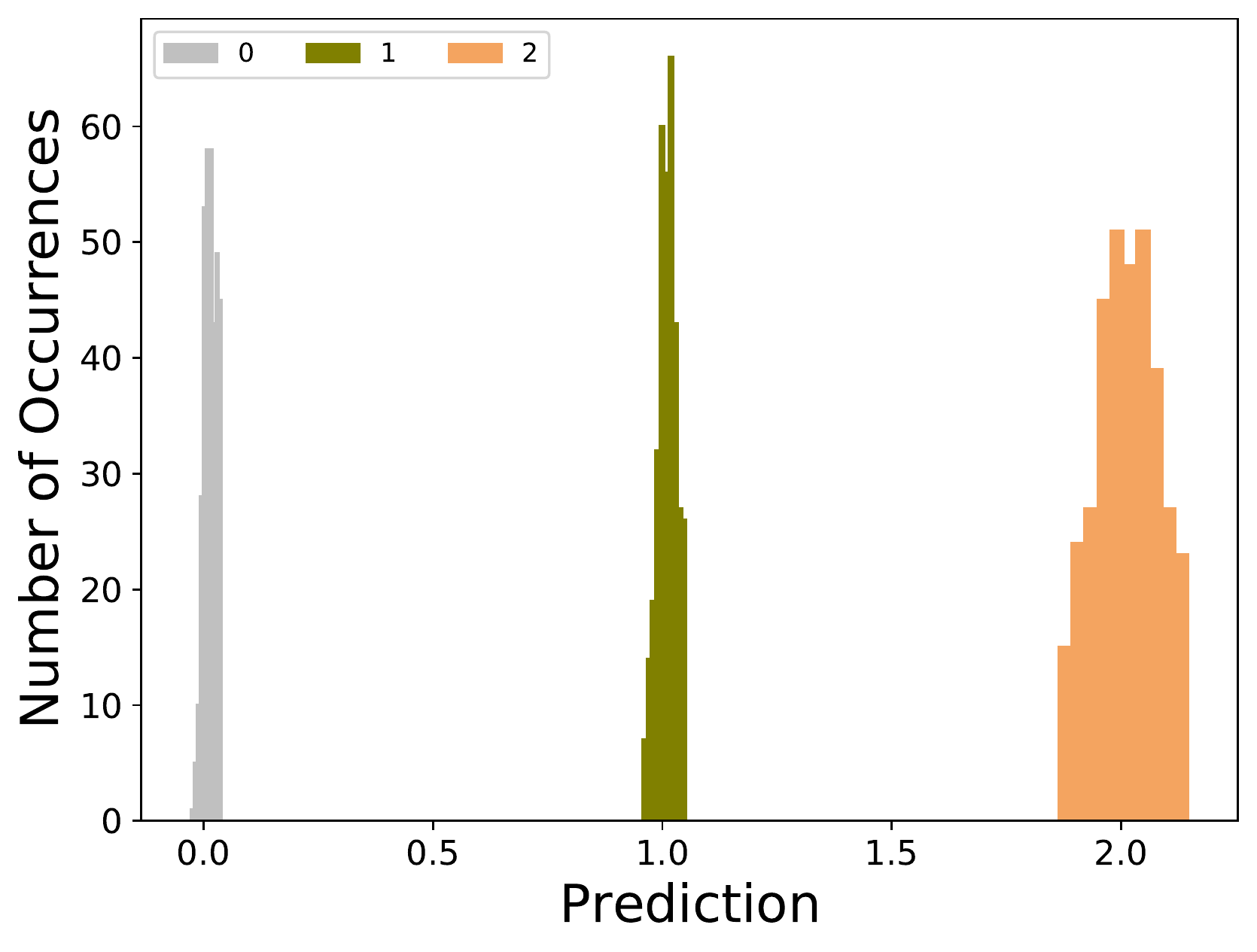}&
   \includegraphics[width=0.17\textwidth]{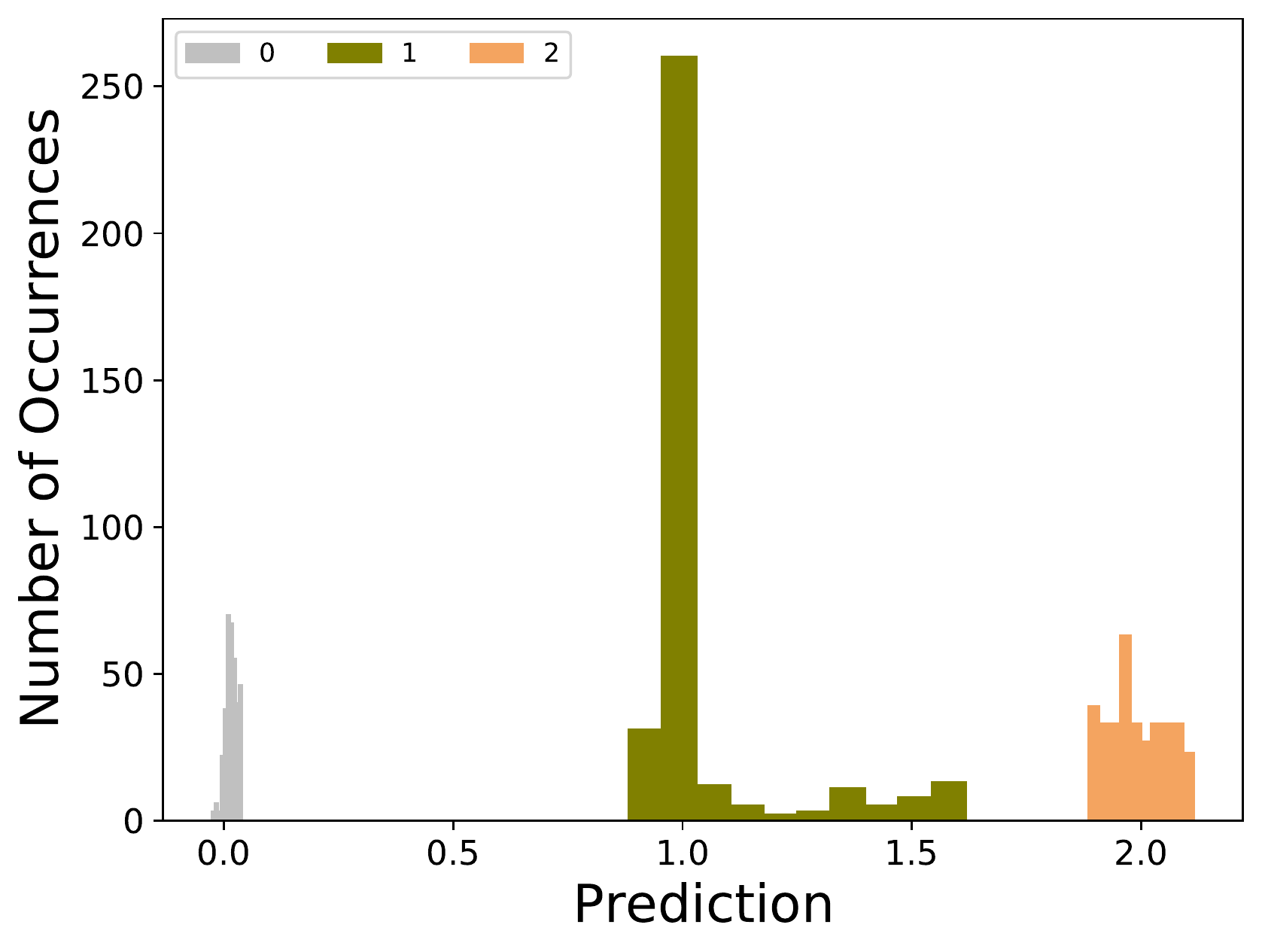}&
   \includegraphics[width=0.17\textwidth]{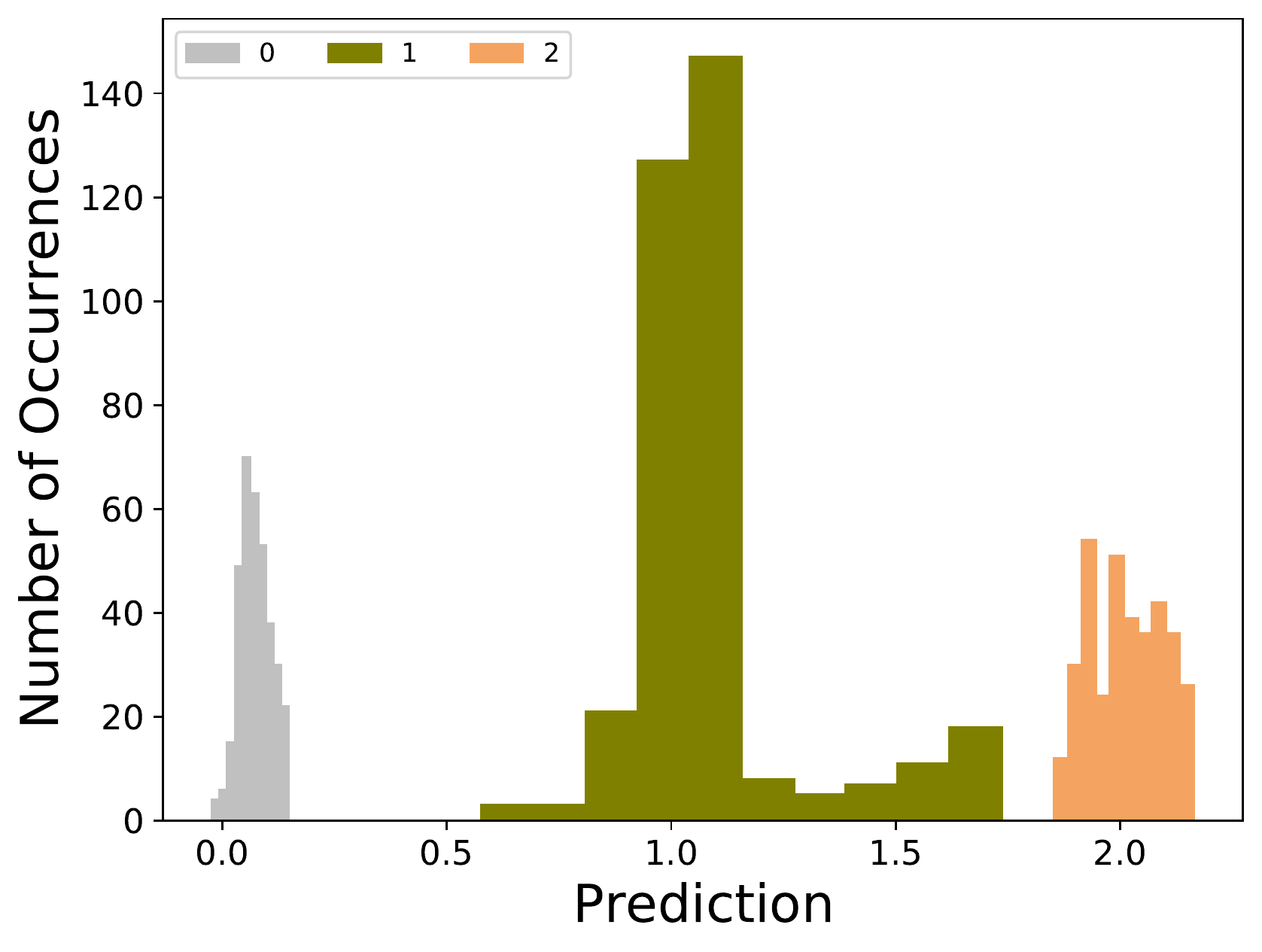}&
   \includegraphics[width=0.17\textwidth]{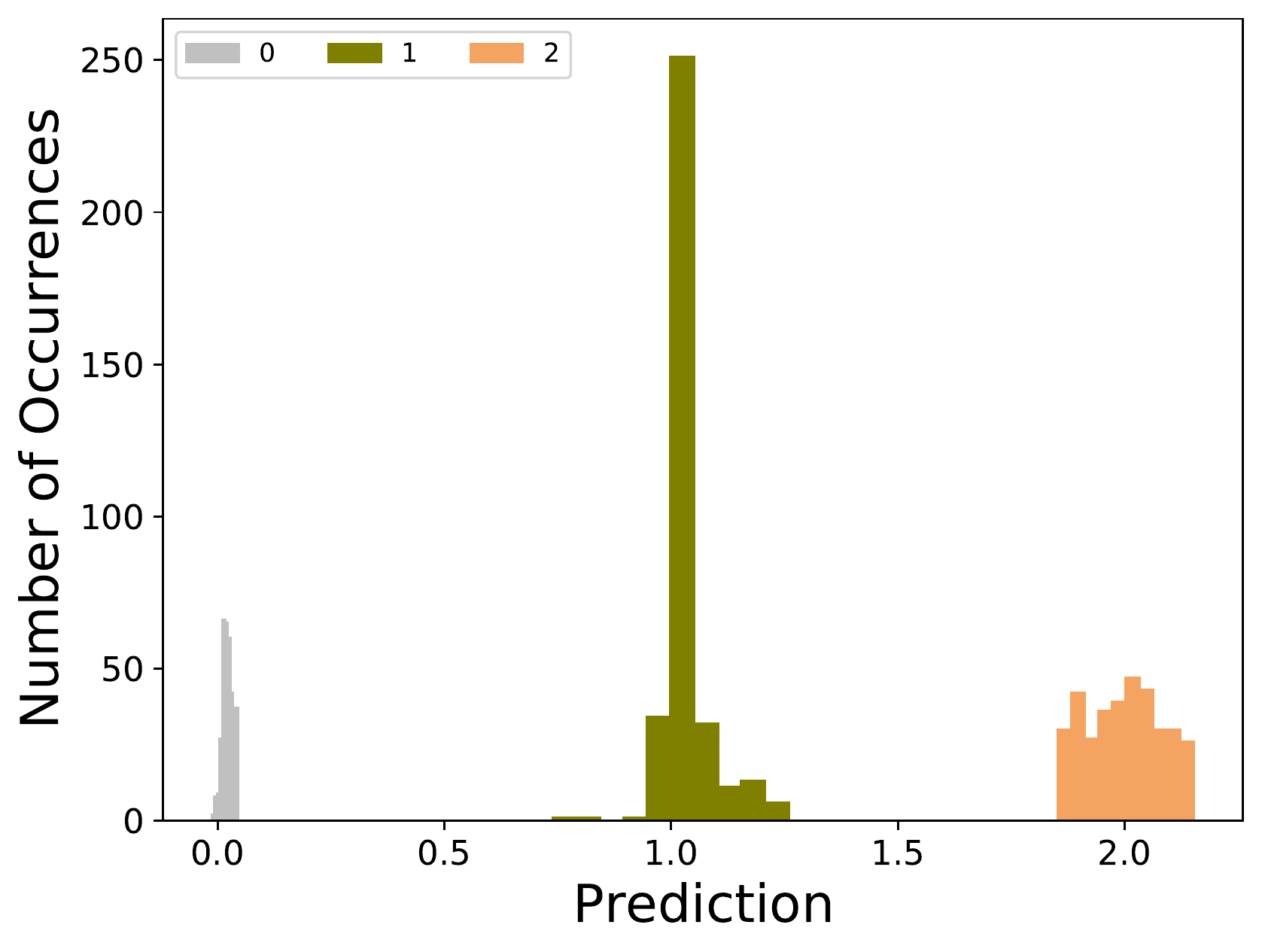} \\
  Digit-0 &Digit-1 &Digit-2 &Digit-3 &Digit-4 \\
   \includegraphics[width=0.17\textwidth]{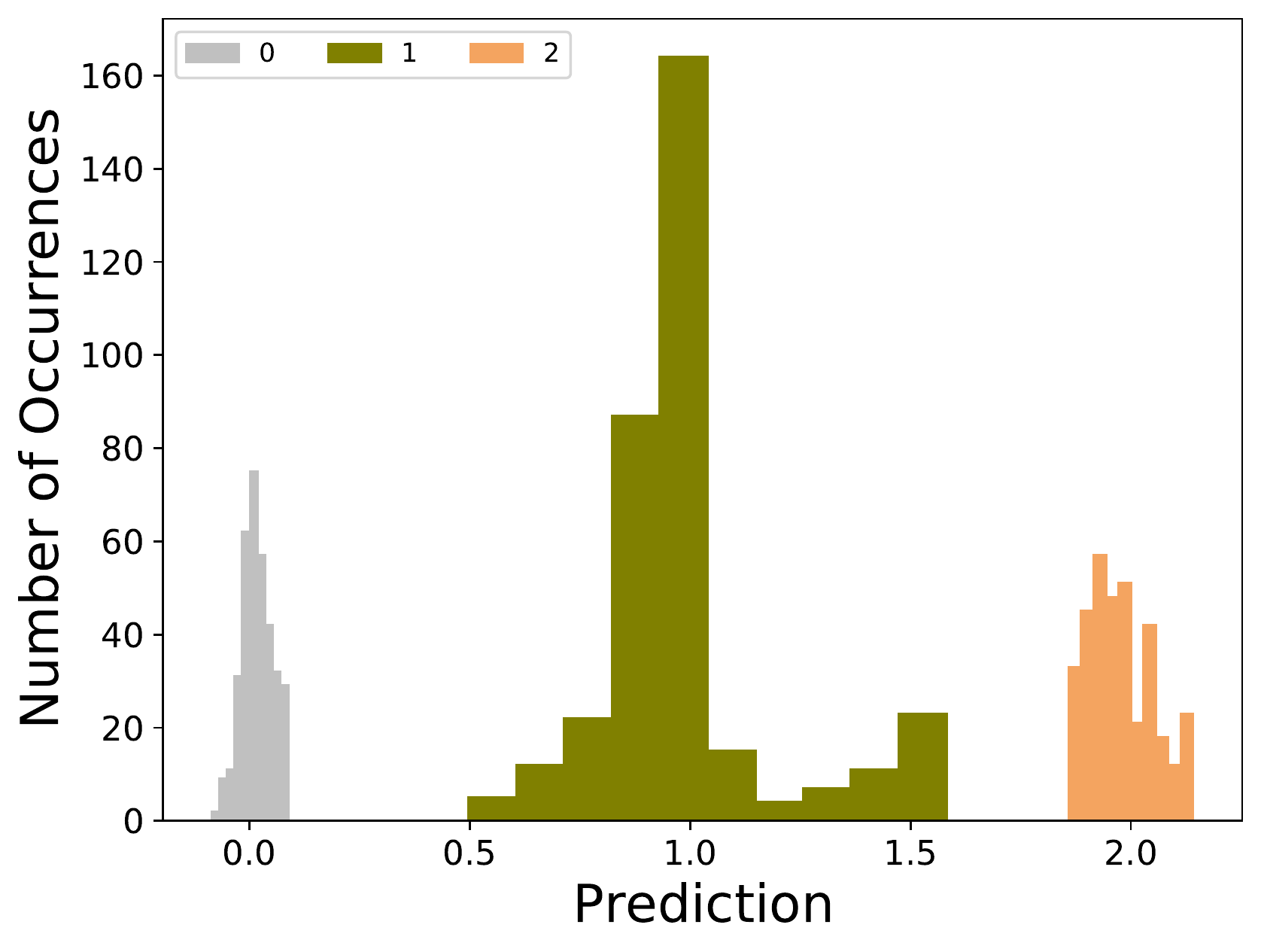}&
   \includegraphics[width=0.17\textwidth]{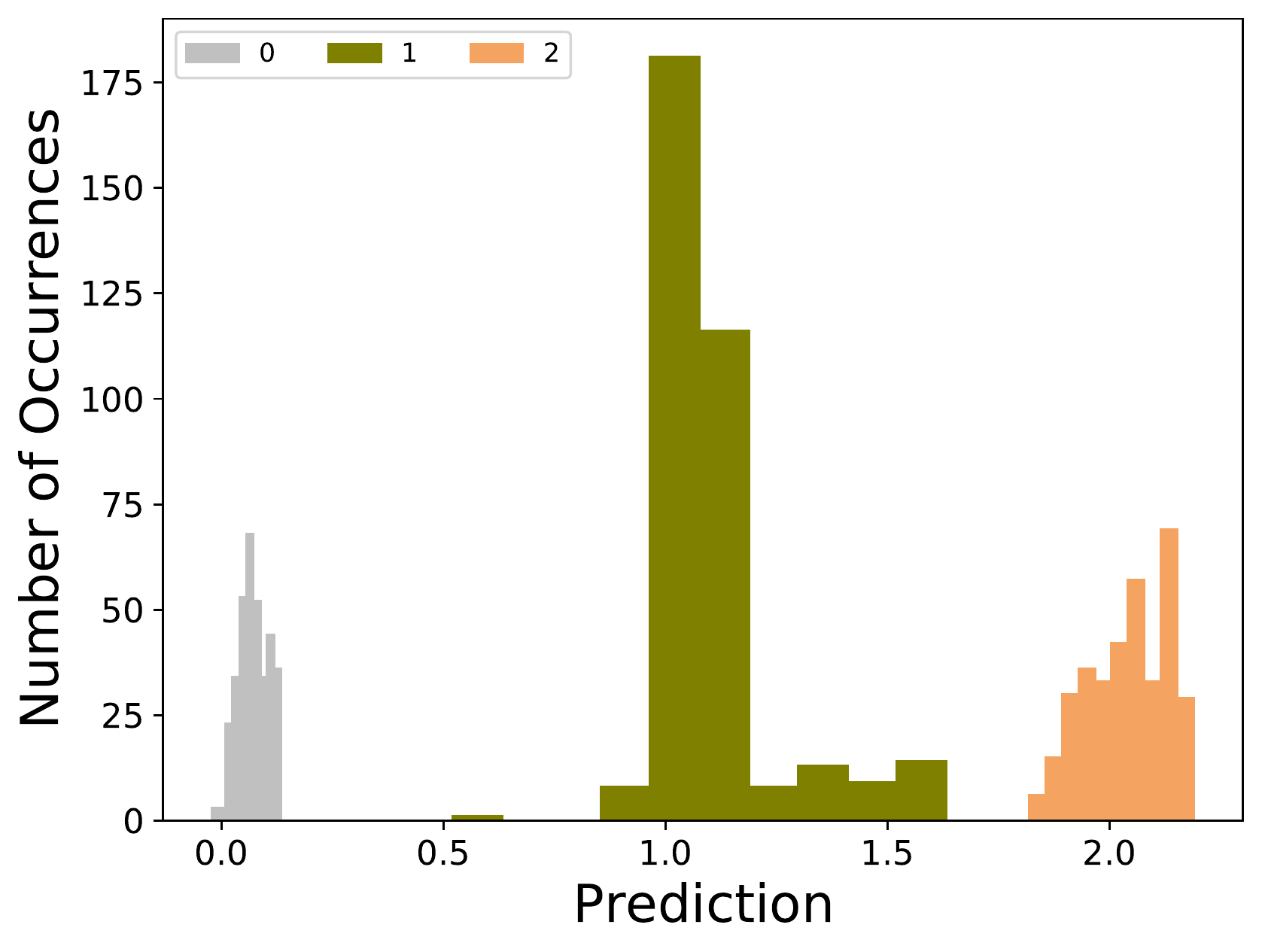}&
   \includegraphics[width=0.17\textwidth]{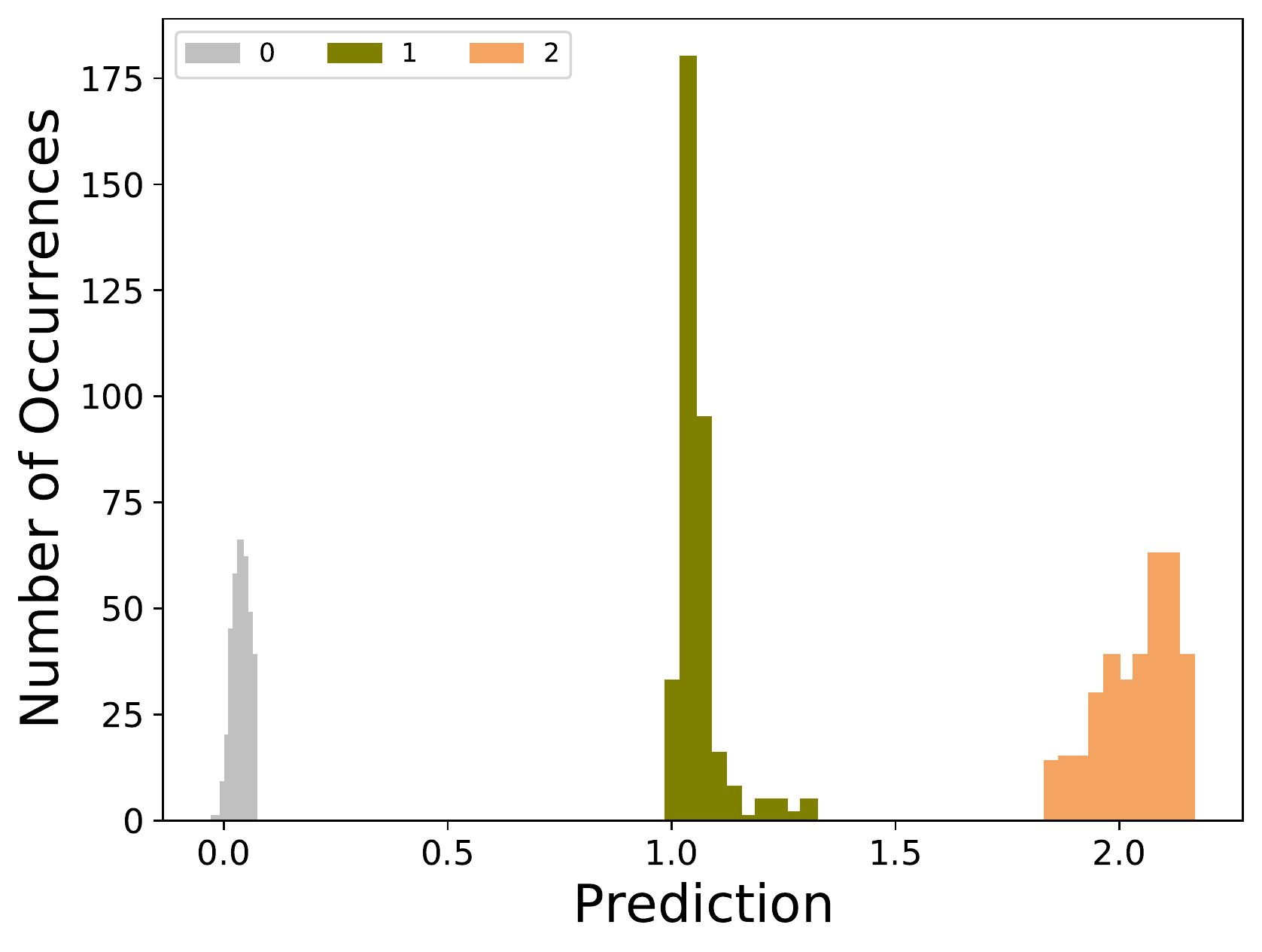}&
   \includegraphics[width=0.17\textwidth]{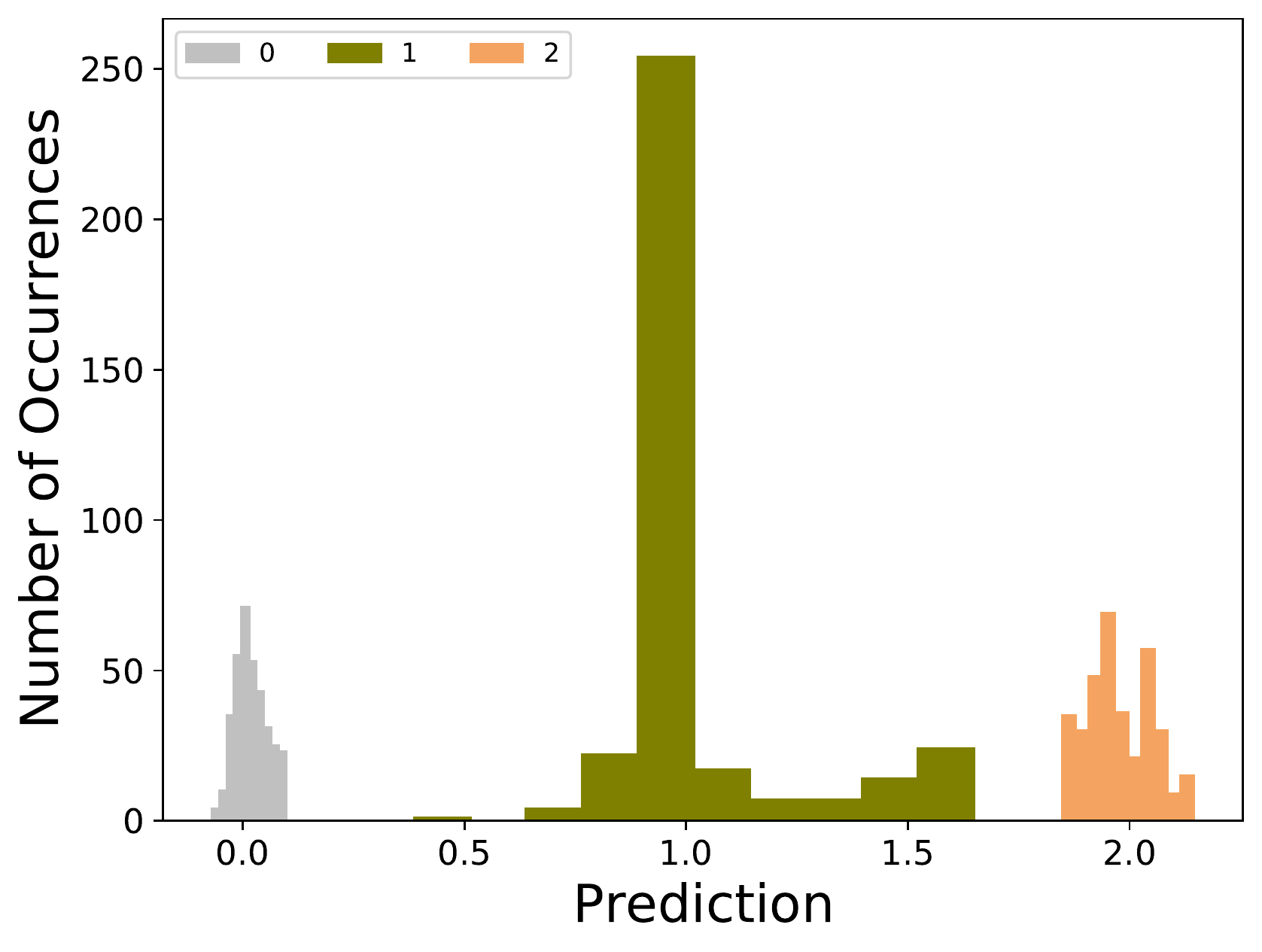}&
   \includegraphics[width=0.17\textwidth]{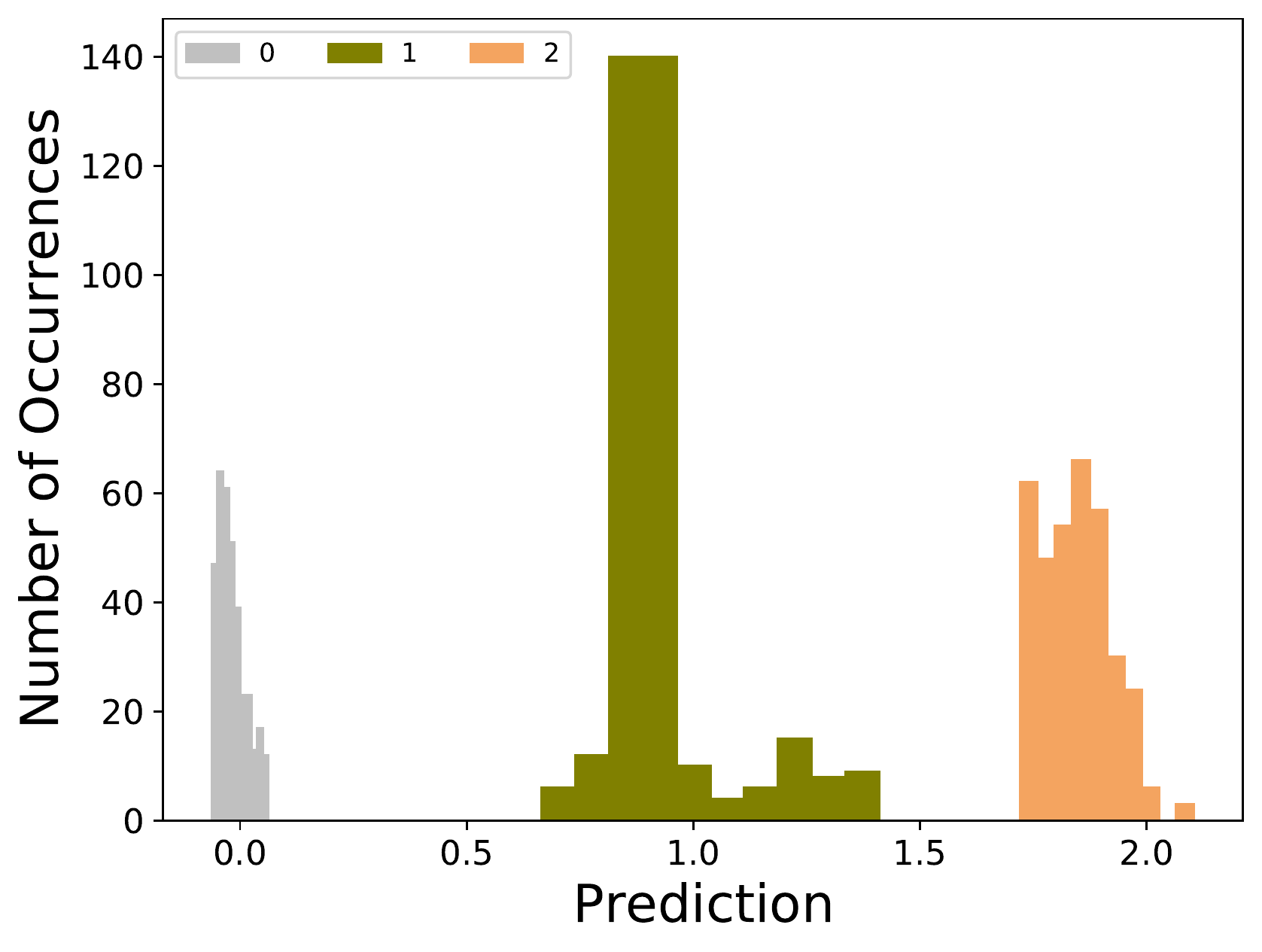} \\
  Digit-5 &Digit-6 &Digit-7 &Digit-8 &Digit-9 
 \end{tabular}}
\caption{Count performance on CLEVR and SVHN images. The figure shows the predicted count values for each count class.} 
\label{fig:count_HIST_SVHNANDCLEVR}
\end{figure*}

The multiple count prediction distribution of cylinder and sphere class of CLEVR-$2$ and that of cylinder, sphere and cube class of CLEVR-$3$ are shown in Figure~\ref{figHISTCLEVR2} and ~\ref{fig:HISTCLEVR3} respectively.
Similarly the multiple count distribution for the ten digit classes in SVHN are visualized in Figure~\ref{fig:HISTSVHN}.
\subsection{Extended qualitative analysis across datasets}
We provide additional qualitative results for CLEVR, SVHN and CityCount images for various count combinations.
\begin{figure*}
 \scriptsize
 \selectfont
 \centering
 \captionsetup[subfigure]{skip=2pt}
  \subcaptionbox{CLEVR-$2$ . \label{subfig:clevr26stylegan}}{
   \begin{tabular}{@{\hspace{.0cm}}c@{\hspace{.02cm}}c@{\hspace{.02cm}}c@{\hspace{.02cm}}c@{\hspace{.05cm}}c@{\hspace{.02cm}}c@{\hspace{.02cm}}c@{\hspace{.02cm}}c@{\hspace{.0cm}}
    }
\includegraphics[width=0.11\textwidth]{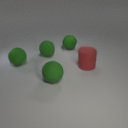} &
\includegraphics[width=0.11\textwidth]{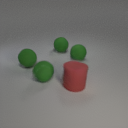} &
\includegraphics[width=0.11\textwidth]{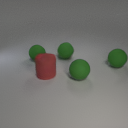} &
\includegraphics[width=0.11\textwidth]{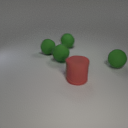} &
 \includegraphics[width=0.11\textwidth]{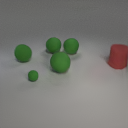} & 
 \includegraphics[width=0.11\textwidth]{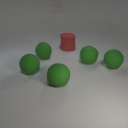} & 
 \includegraphics[width=0.11\textwidth]{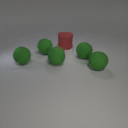} & 
 \includegraphics[width=0.11\textwidth]{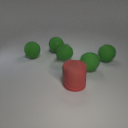} 
 \\
 \multicolumn{4}{@{\hspace{.01cm}}c@{\hspace{.01cm}}}{{[}1 4]} &
 \multicolumn{4}{@{\hspace{.01cm}}c@{\hspace{.01cm}}}{{[}1 5]} \\
\includegraphics[width=0.11\textwidth]{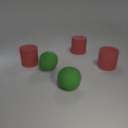} & 
\includegraphics[width=0.11\textwidth]{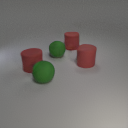} & 
\includegraphics[width=0.11\textwidth]{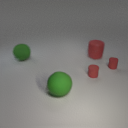} &  
\includegraphics[width=0.11\textwidth]{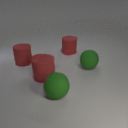} & 
 \includegraphics[width=0.11\textwidth]{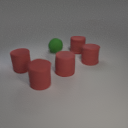} & 
 \includegraphics[width=0.11\textwidth]{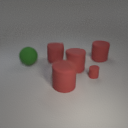} &  
 \includegraphics[width=0.11\textwidth]{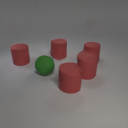} &  
 \includegraphics[width=0.11\textwidth]{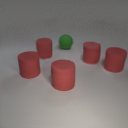} 
 \\
 \multicolumn{4}{@{\hspace{.01cm}}c@{\hspace{.01cm}}}{{[}3 2]} &
 \multicolumn{4}{@{\hspace{.01cm}}c@{\hspace{.01cm}}}{{[}5 1]} \\
 \end{tabular}}
 \subcaptionbox{CLEVR-$3$. \label{subfig:clevr3stylegan}}{
   \begin{tabular}{@{\hspace{.0cm}}c@{\hspace{.02cm}}c@{\hspace{.02cm}}c@{\hspace{.02cm}}c@{\hspace{.05cm}}c@{\hspace{.02cm}}c@{\hspace{.02cm}}c@{\hspace{.02cm}}c@{\hspace{.0cm}}
    }
\includegraphics[width=0.11\textwidth]{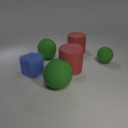} & 
\includegraphics[width=0.11\textwidth]{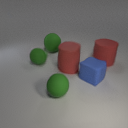} & 
\includegraphics[width=0.11\textwidth]{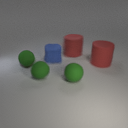} & 
\includegraphics[width=0.11\textwidth]{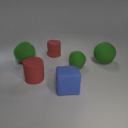} & 
 \includegraphics[width=0.11\textwidth]{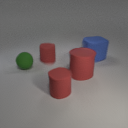} & 
\includegraphics[width=0.11\textwidth]{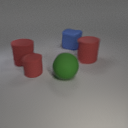} & 
\includegraphics[width=0.11\textwidth]{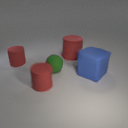} & 
\includegraphics[width=0.11\textwidth]{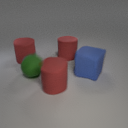} 
 \\
 \multicolumn{4}{@{\hspace{.01cm}}c@{\hspace{.01cm}}}{{[}2 3 1]} &
 \multicolumn{4}{@{\hspace{.01cm}}c@{\hspace{.01cm}}}{{[}3 1 1]} \\
\includegraphics[width=0.11\textwidth]{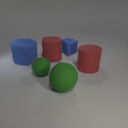} & 
\includegraphics[width=0.11\textwidth]{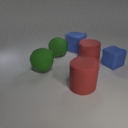} & 
\includegraphics[width=0.11\textwidth]{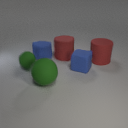} & 
\includegraphics[width=0.11\textwidth]{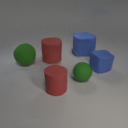} & 
 \includegraphics[width=0.11\textwidth]{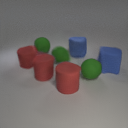} & 
 \includegraphics[width=0.11\textwidth]{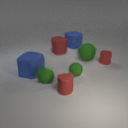} & 
  \includegraphics[width=0.11\textwidth]{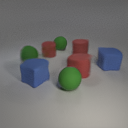} & 
   \includegraphics[width=0.11\textwidth]{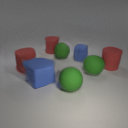} 
 \\
 \multicolumn{4}{@{\hspace{.01cm}}c@{\hspace{.01cm}}}{{[}2 2 2]} &
 \multicolumn{4}{@{\hspace{.01cm}}c@{\hspace{.01cm}}}{{[}3 3 2]} \\
 \end{tabular}}
 \subcaptionbox{CLEVR-$2$ multicolor. \label{subfig:clev6rmulticolorstylegan}}{
    \begin{tabular}{@{\hspace{.0cm}}c@{\hspace{.02cm}}c@{\hspace{.02cm}}c@{\hspace{.02cm}}c@{\hspace{.05cm}}c@{\hspace{.02cm}}c@{\hspace{.02cm}}c@{\hspace{.02cm}}c@{\hspace{.0cm}}
    }
\includegraphics[width=0.11\textwidth]{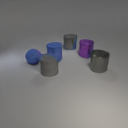} & 
 \includegraphics[width=0.11\textwidth]{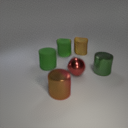} &
 \includegraphics[width=0.11\textwidth]{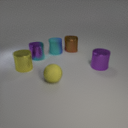} &
 \includegraphics[width=0.11\textwidth]{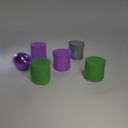} &
 \includegraphics[width=0.11\textwidth]{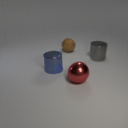} &
\includegraphics[width=0.11\textwidth]{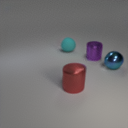} &
\includegraphics[width=0.11\textwidth]{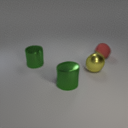} &
\includegraphics[width=0.11\textwidth]{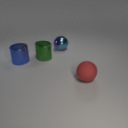} 
 \\
 \multicolumn{4}{@{\hspace{.01cm}}c@{\hspace{.01cm}}}{{[}5 1]} &
 \multicolumn{4}{@{\hspace{.01cm}}c@{\hspace{.01cm}}}{{[}2 2]} \\
\includegraphics[width=0.11\textwidth]{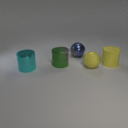} & 
\includegraphics[width=0.11\textwidth]{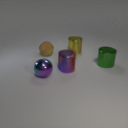} & 
\includegraphics[width=0.11\textwidth]{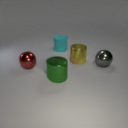} & 
\includegraphics[width=0.11\textwidth]{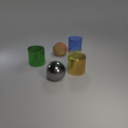} & 
\includegraphics[width=0.11\textwidth]{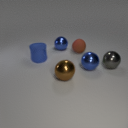} & 
\includegraphics[width=0.11\textwidth]{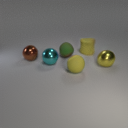} & 
\includegraphics[width=0.11\textwidth]{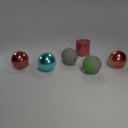} &  
\includegraphics[width=0.11\textwidth]{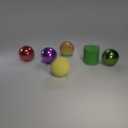} 
 \\
 \multicolumn{4}{@{\hspace{.01cm}}c@{\hspace{.01cm}}}{{[}3 2]} &
 \multicolumn{4}{@{\hspace{.01cm}}c@{\hspace{.01cm}}}{{[}1 5]} \\
 \end{tabular}}
 \subcaptionbox{CLEVR-$3$ multicolor. \label{subfig:clevr2multicolorstylegan}}{
    \begin{tabular}{@{\hspace{.0cm}}c@{\hspace{.02cm}}c@{\hspace{.02cm}}c@{\hspace{.02cm}}c@{\hspace{.05cm}}c@{\hspace{.02cm}}c@{\hspace{.02cm}}c@{\hspace{.02cm}}c@{\hspace{.0cm}}
    }
\includegraphics[width=0.11\textwidth]{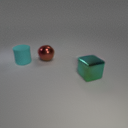} & 
\includegraphics[width=0.11\textwidth]{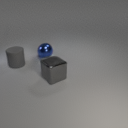} & 
\includegraphics[width=0.11\textwidth]{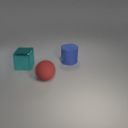} & 
\includegraphics[width=0.11\textwidth]{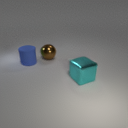} & 
 \includegraphics[width=0.11\textwidth]{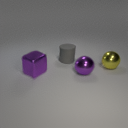}& 
 \includegraphics[width=0.11\textwidth]{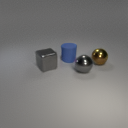} & 
  \includegraphics[width=0.11\textwidth]{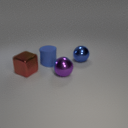} & 
   \includegraphics[width=0.11\textwidth]{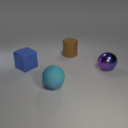}
 \\
 \multicolumn{4}{@{\hspace{.01cm}}c@{\hspace{.01cm}}}{{[}1 1 1]} &
 \multicolumn{4}{@{\hspace{.01cm}}c@{\hspace{.01cm}}}{{[}1 2 1]} \\
\includegraphics[width=0.11\textwidth]{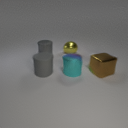} &
\includegraphics[width=0.11\textwidth]{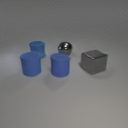} & 
\includegraphics[width=0.11\textwidth]{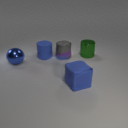} & 
\includegraphics[width=0.11\textwidth]{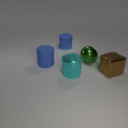} & 
 \includegraphics[width=0.11\textwidth]{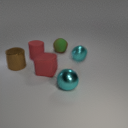}& 
 \includegraphics[width=0.11\textwidth]{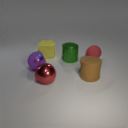} & 
  \includegraphics[width=0.11\textwidth]{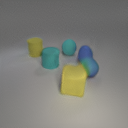} & 
   \includegraphics[width=0.11\textwidth]{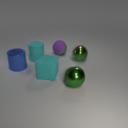} 
 \\
 \multicolumn{4}{@{\hspace{.01cm}}c@{\hspace{.01cm}}}{{[}3 1 1]} &
 \multicolumn{4}{@{\hspace{.01cm}}c@{\hspace{.01cm}}}{{[}2 3 1]} \\
 \end{tabular}}
 \caption{Generated MC$^2$-StyleGAN2 CLEVR images
 for different count combinations. The result shows better quality images with correct count condition.
 }\label{fig:clevr_images_stylegan2}
 \end{figure*}
 The generated images for CLEVR-$2$ and CLEVR-$3$, both with same color and multicolor for similar shapes are shown in Figure~\ref{fig:clevr_images_stylegan2}.
 \begin{figure*}
 \scriptsize
 \selectfont
 \centering
  \subcaptionbox{Generated SVHN-$2$ images. \label{fig:svhn_generatedstylegan}}{
    \begin{tabular}{@{\hspace{.0cm}}c@{\hspace{.04cm}}c@{\hspace{.04cm}}c@{\hspace{.04cm}}c@{\hspace{.04cm}}c@{\hspace{.04cm}}c@{\hspace{.04cm}}c@{\hspace{.04cm}}c@{\hspace{.0cm}}
    }
\includegraphics[width=0.101\textwidth]{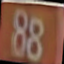} & 
 \includegraphics[width=0.101\textwidth]{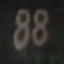} &
 \includegraphics[width=0.101\textwidth]{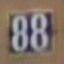} &
 \includegraphics[width=0.101\textwidth]{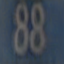} &
 \includegraphics[width=0.101\textwidth]{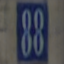} &
 \includegraphics[width=0.101\textwidth]{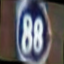} &
 \includegraphics[width=0.101\textwidth]{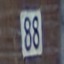} &
 \includegraphics[width=0.101\textwidth]{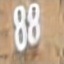} 
 \\
 \multicolumn{8}{@{\hspace{.01cm}}c@{\hspace{.01cm}}}{{[}0 0 0 0 0 0 0 0 2 0]} \\
\\
\includegraphics[width=0.101\textwidth]{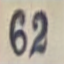} & 
 \includegraphics[width=0.101\textwidth]{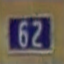} &
\includegraphics[width=0.101\textwidth]{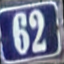} & 
\includegraphics[width=0.101\textwidth]{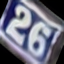} &
\includegraphics[width=0.101\textwidth]{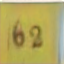} & 
\includegraphics[width=0.101\textwidth]{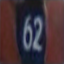} & 
\includegraphics[width=0.101\textwidth]{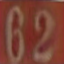} & \includegraphics[width=0.101\textwidth]{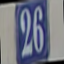} 
 \\
  \multicolumn{8}{@{\hspace{.01cm}}c@{\hspace{.01cm}}}{{[}0 0 1 0 0 0 1 0 0 0]}  \\
\\
\includegraphics[width=0.101\textwidth]{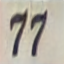} & \includegraphics[width=0.101\textwidth]{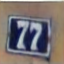} &
\includegraphics[width=0.101\textwidth]{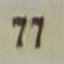} &
\includegraphics[width=0.101\textwidth]{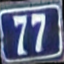} &
\includegraphics[width=0.101\textwidth]{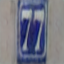} &
\includegraphics[width=0.101\textwidth]{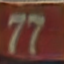} &
\includegraphics[width=0.101\textwidth]{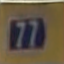} &
\includegraphics[width=0.101\textwidth]{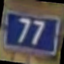} 
 \\
 \multicolumn{8}{@{\hspace{.01cm}}c@{\hspace{.01cm}}}{{[}0 0 0 0 0 0 0 2 0 0]}  \\
 \\
\includegraphics[width=0.101\textwidth]{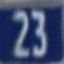} & 
\includegraphics[width=0.101\textwidth]{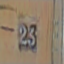} & 
\includegraphics[width=0.101\textwidth]{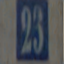} & 
\includegraphics[width=0.101\textwidth]{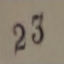} & 
\includegraphics[width=0.101\textwidth]{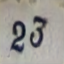} & 
\includegraphics[width=0.101\textwidth]{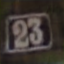} & 
\includegraphics[width=0.101\textwidth]{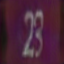} & 
\includegraphics[width=0.101\textwidth]{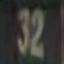} 
 \\
 \multicolumn{8}{@{\hspace{.01cm}}c@{\hspace{.01cm}}}{{[}0 0 1 1 0 0 0 0 0 0]}  \\
\\
\includegraphics[width=0.101\textwidth]{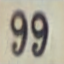} & \includegraphics[width=0.101\textwidth]{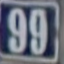} &
\includegraphics[width=0.101\textwidth]{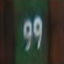} &
\includegraphics[width=0.101\textwidth]{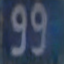} &
\includegraphics[width=0.101\textwidth]{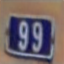} &
\includegraphics[width=0.101\textwidth]{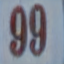} &
\includegraphics[width=0.101\textwidth]{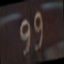} &
\includegraphics[width=0.101\textwidth]{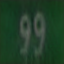} 
 \\
 \multicolumn{8}{@{\hspace{.01cm}}c@{\hspace{.01cm}}}{{[}0 0 0 0 0 0 0 0 0 2]}  \\
 \\
\includegraphics[width=0.101\textwidth]{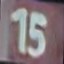} & 
\includegraphics[width=0.101\textwidth]{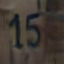} & 
\includegraphics[width=0.101\textwidth]{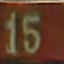} & 
\includegraphics[width=0.101\textwidth]{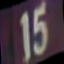} & 
\includegraphics[width=0.101\textwidth]{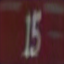} & 
\includegraphics[width=0.101\textwidth]{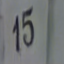} & 
\includegraphics[width=0.101\textwidth]{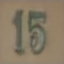} & 
\includegraphics[width=0.101\textwidth]{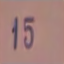} 
 \\
\multicolumn{8}{@{\hspace{.01cm}}c@{\hspace{.01cm}}}{{[}0 1 0 0 0 1 0 0 0 0]}  \\
\\
\includegraphics[width=0.101\textwidth]{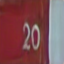} &
\includegraphics[width=0.101\textwidth]{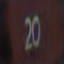} &
\includegraphics[width=0.101\textwidth]{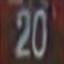} &
\includegraphics[width=0.101\textwidth]{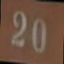} &
\includegraphics[width=0.101\textwidth]{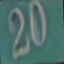} &
\includegraphics[width=0.101\textwidth]{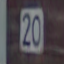} &
\includegraphics[width=0.101\textwidth]{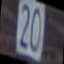} &
\includegraphics[width=0.101\textwidth]{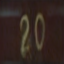} 
 \\
\multicolumn{8}{@{\hspace{.01cm}}c@{\hspace{.01cm}}}{{[}1 0 1 0 0 0 0 0 0 0]}  \\
\\
\includegraphics[width=0.101\textwidth]{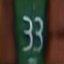} & 
\includegraphics[width=0.101\textwidth]{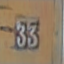} & 
\includegraphics[width=0.101\textwidth]{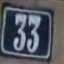} & 
\includegraphics[width=0.101\textwidth]{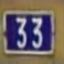} & 
\includegraphics[width=0.101\textwidth]{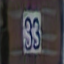} & 
\includegraphics[width=0.101\textwidth]{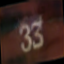} & 
\includegraphics[width=0.101\textwidth]{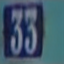} & 
\includegraphics[width=0.101\textwidth]{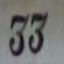} 
 \\
\multicolumn{8}{@{\hspace{.01cm}}c@{\hspace{.01cm}}}{{[}0 0 0 2 0 0 0 0 0 0]}  \\
\\
\includegraphics[width=0.101\textwidth]{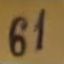} &
\includegraphics[width=0.101\textwidth]{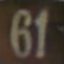} &
\includegraphics[width=0.101\textwidth]{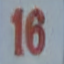} &
\includegraphics[width=0.101\textwidth]{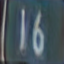} &
\includegraphics[width=0.101\textwidth]{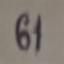} &
\includegraphics[width=0.101\textwidth]{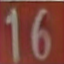} &
\includegraphics[width=0.101\textwidth]{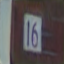} &
\includegraphics[width=0.101\textwidth]{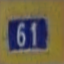} 
 \\
\multicolumn{8}{@{\hspace{.01cm}}c@{\hspace{.01cm}}}{{[}0 1 0 0 0 0 1 0 0 0]}  \\
\end{tabular}}
 \caption{Generated MC$^2$-StyleGAN2 SVHN-$2$ images 
 for different ten-dimensional count vectors. The generated images are of better image quality with correct number of object instances according to the condition.}\label{fig:SVHNSTYLEGAN2}
 \end{figure*}
 The additional results for SVHN images along with their respective input count are shown in Figure \ref{fig:SVHNSTYLEGAN2}.
 \begin{figure*}
 \scriptsize
 \selectfont
 \centering
\subcaptionbox{Real and Generated CityCount images. \label{fig:citycount_generated}}{
 \begin{tabular}{@{\hspace{.0cm}}c@{\hspace{.1cm}}c@{\hspace{.03cm}}c@{\hspace{.03cm}}c@{\hspace{.03cm}}c@{\hspace{.03cm}}c@{\hspace{.03cm}}c@{\hspace{.0cm}}
    }
     \includegraphics[width=0.14\textwidth]{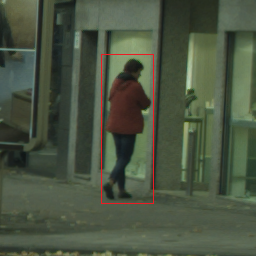} &
     \includegraphics[width=0.14\textwidth]{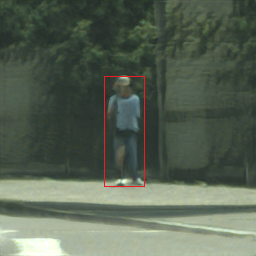} & 
     \includegraphics[width=0.14\textwidth]{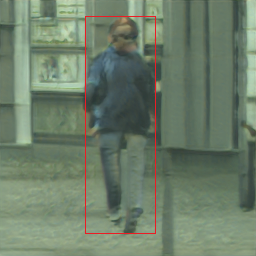} &
     \includegraphics[width=0.14\textwidth]{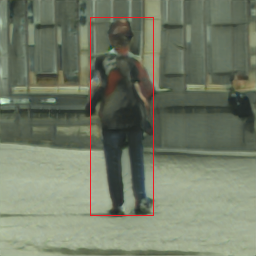}&
      \includegraphics[width=0.14\textwidth]{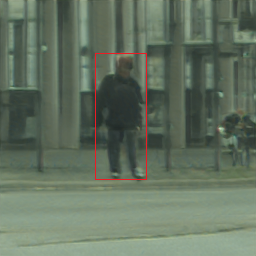}&
      \includegraphics[width=0.14\textwidth]{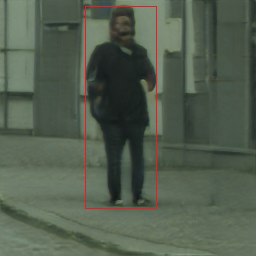}&
 \includegraphics[width=0.14\textwidth]{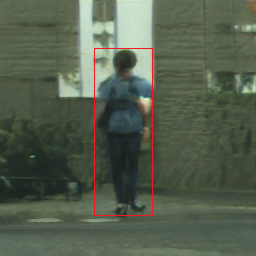} \\
     real & \multicolumn{6}{@{\hspace{.01cm}}c@{\hspace{.01cm}}}{{[}0 1]}  \\
     \includegraphics[width=0.14\textwidth]{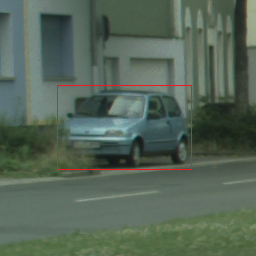} &
     \includegraphics[width=0.14\textwidth]{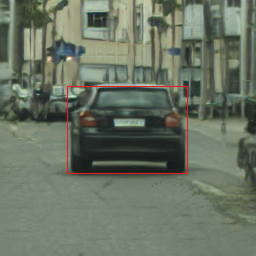} &
     \includegraphics[width=0.14\textwidth]{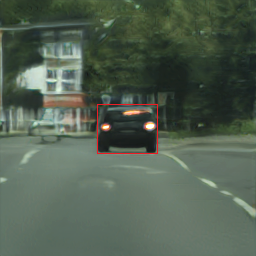} &
     \includegraphics[width=0.14\textwidth]{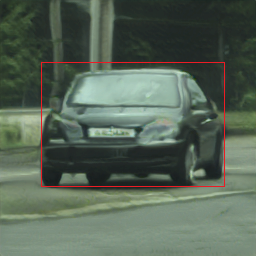}&
      \includegraphics[width=0.14\textwidth]{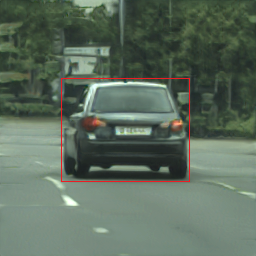}&
      \includegraphics[width=0.14\textwidth]{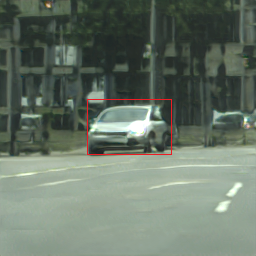}&
 \includegraphics[width=0.14\textwidth]{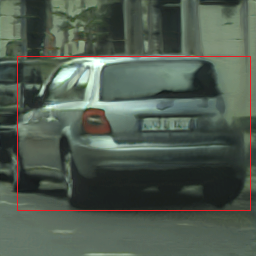}\\
     real & \multicolumn{6}{@{\hspace{.01cm}}c@{\hspace{.01cm}}}{{[}1 0]}  \\
     \includegraphics[width=0.14\textwidth]{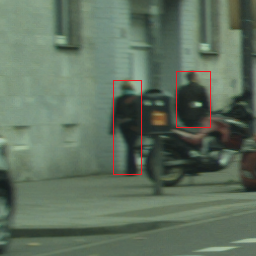} &
     \includegraphics[width=0.14\textwidth]{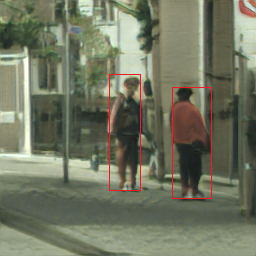} &
     \includegraphics[width=0.14\textwidth]{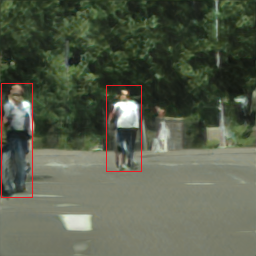} &
         \includegraphics[width=0.14\textwidth]{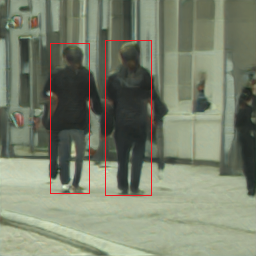} &
      \includegraphics[width=0.14\textwidth]{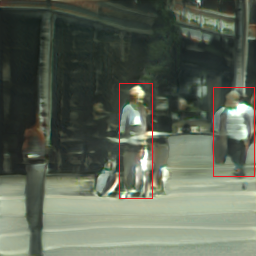}&
      \includegraphics[width=0.14\textwidth]{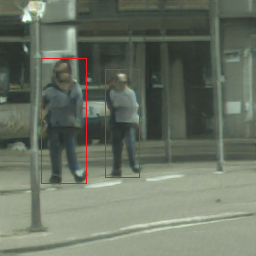} &
     \includegraphics[width=0.14\textwidth]{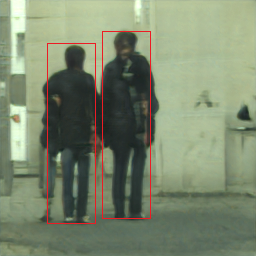} \\           
      
     real & \multicolumn{6}{@{\hspace{.01cm}}c@{\hspace{.01cm}}}{{[}0 2]}  \\
     \includegraphics[width=0.14\textwidth]{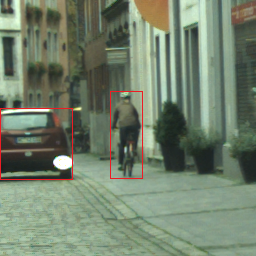} &
     \includegraphics[width=0.14\textwidth]{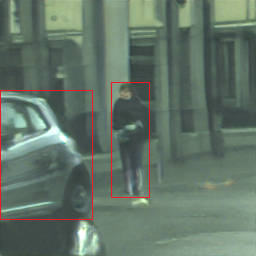}&
     \includegraphics[width=0.14\textwidth]{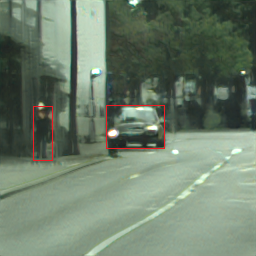} &
      \includegraphics[width=0.14\textwidth]{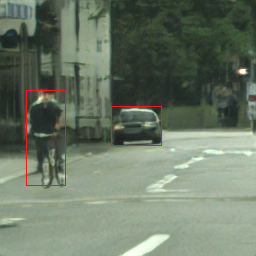} &
       \includegraphics[width=0.14\textwidth]{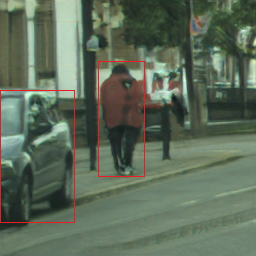} &
      \includegraphics[width=0.14\textwidth]{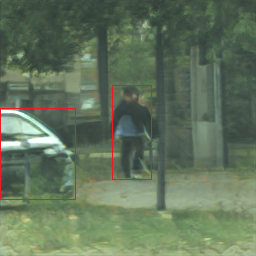}&
 \includegraphics[width=0.14\textwidth]{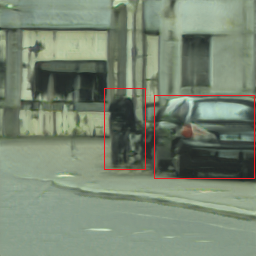}\\
     real & \multicolumn{6}{@{\hspace{.01cm}}c@{\hspace{.01cm}}}{{[}1 1]}  \\
     \includegraphics[width=0.14\textwidth]{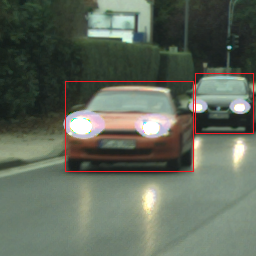} &
     \includegraphics[width=0.14\textwidth]{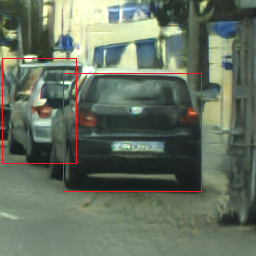} &
     \includegraphics[width=0.14\textwidth]{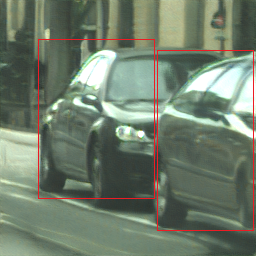} &
     \includegraphics[width=0.14\textwidth]{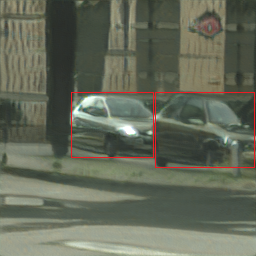} &
      \includegraphics[width=0.14\textwidth]{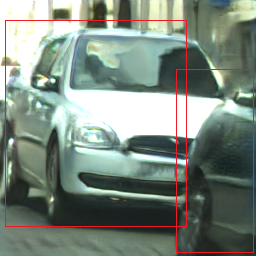}&
      \includegraphics[width=0.14\textwidth]{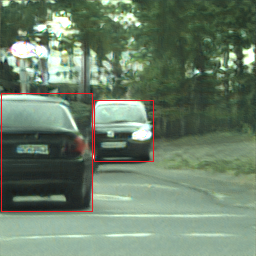}&
 \includegraphics[width=0.14\textwidth]{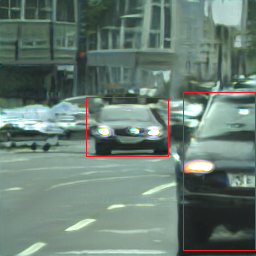}\\
     real & \multicolumn{6}{@{\hspace{.01cm}}c@{\hspace{.01cm}}}{{[}2 0]}  \\
     \includegraphics[width=0.14\textwidth]{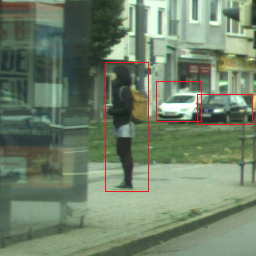} &
     \includegraphics[width=0.14\textwidth]{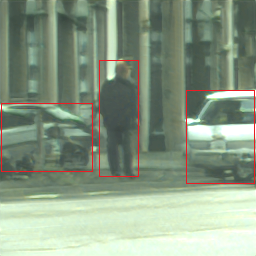}& 
     \includegraphics[width=0.14\textwidth]{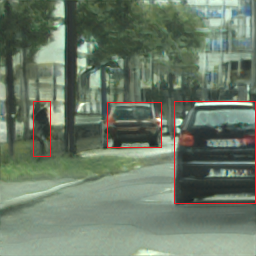} &
     \includegraphics[width=0.14\textwidth]{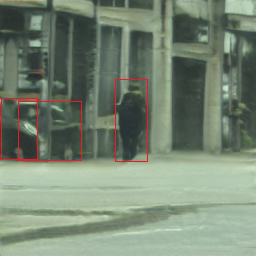}&
      \includegraphics[width=0.14\textwidth]{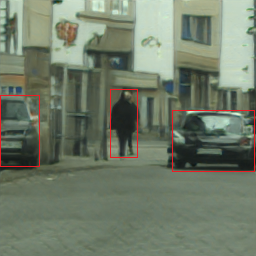}&
      \includegraphics[width=0.14\textwidth]{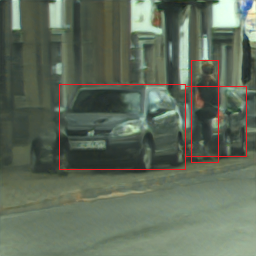}&
 \includegraphics[width=0.14\textwidth]{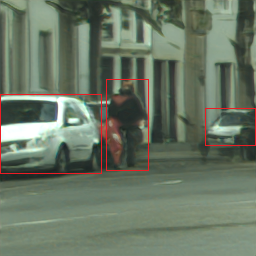} \\
     real & \multicolumn{6}{@{\hspace{.01cm}}c@{\hspace{.01cm}}}{{[}2 1]}  \\
    \includegraphics[width=0.14\textwidth]{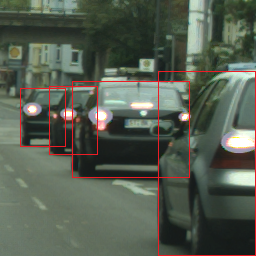} & 
    \includegraphics[width=0.14\textwidth]{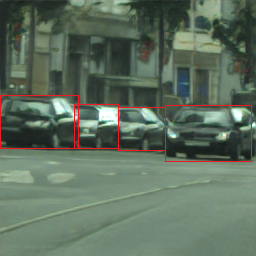} &
     \includegraphics[width=0.14\textwidth]{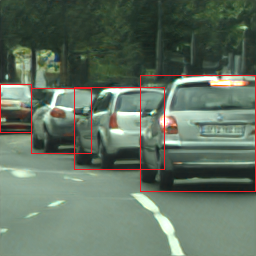} &
    \includegraphics[width=0.14\textwidth]{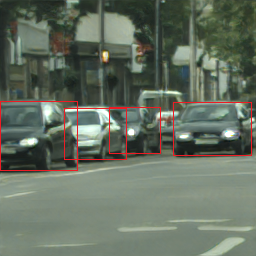} &
    \includegraphics[width=0.14\textwidth]{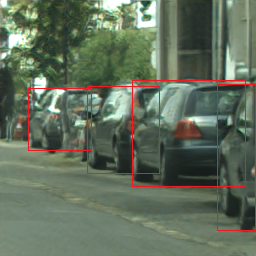} &
    \includegraphics[width=0.14\textwidth]{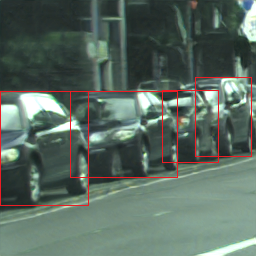} &
 \includegraphics[width=0.14\textwidth]{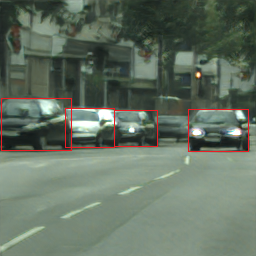}  \\
     real & \multicolumn{6}{@{\hspace{.01cm}}c@{\hspace{.01cm}}}{{[}4 0]}  \\
    \end{tabular}}
 \caption{Real and generated CityCount images by our MC$^2$-StyleGAN2. The model generates good quality images with correct number of cars/persons.}
 \label{citycountstylegan2}
 \end{figure*}
The extended results for CityCount images along with their respective input count are shown in Figure \ref{citycountstylegan2}.
For the ease of visualization boxes are drawn around objects of interest.
The generated images exhibit diversity and good quality (FID values are given in the main paper) across all the datasets considered.
\section{Network architecture and implementation details}
\subsection{MC$^2$-SimpleGAN}
\begin{figure*}[!htbp] 
\hspace{3.25cm}
\begin{minipage}[l]{0.67\textwidth}
    \begin{subfigure}{\textwidth}
         \centering
         \includegraphics[width=0.66\textwidth]{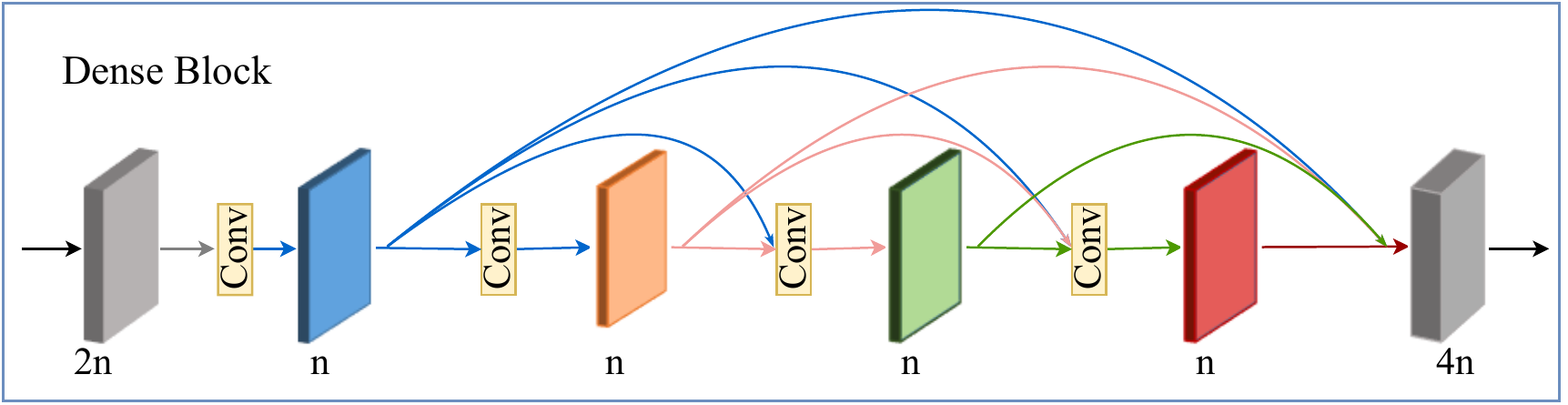}
         \caption{Dense block: Three layer dense block with growth rate n}
         \label{denseblock}\vspace{1.1cm}
     \end{subfigure}
    \begin{subfigure}{\textwidth}
         \includegraphics[width=0.99\textwidth]{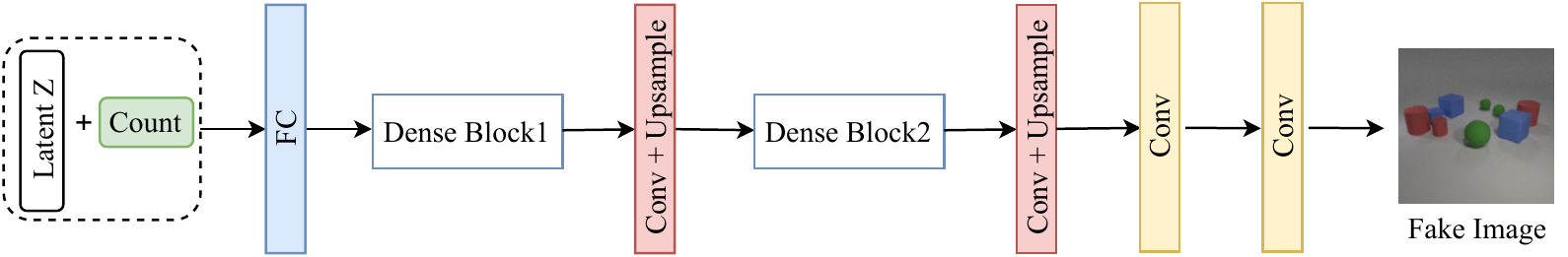}
         \caption{MC$^2$-SimpleGAN Generator: A noise sample and a multi-class count vector are passed to a fully connected layer (FC). Two dense blocks (details see above) coupled with conv.~and upsampling layers are followed by two conv.~layers.
         }
         \label{genblock}
     \end{subfigure}
     \end{minipage}
    \caption{MC$^2$-SimpleGAN Generator.}
    \label{MCGANGen}
\end{figure*}
We introduce MC$^2$-SimpleGAN as a simple network based architecture of our proposed method for fair and easy comparison study with other conditional GAN variants.
The generator of the MC$^2$-SimpleGAN is shown in Figure~\ref{MCGANGen}.
The architecture comprises of a count conditioned generator and a discriminator , that is equipped with an additional count prediction network.
Our basic generator network is inspired from the DenseNet architecture. 
DenseNet introduces dense blocks which consist of several convolutional layers where the output from each layer is connected in a feed forward fashion to its succeeding layers (see Figure~\ref{denseblock} for a visualization). 
The additional skip connections in the dense blocks strengthen the count conditioning in the generator since the input feature maps are connected to the output layers of the dense block.
We use two dense blocks of three layers with a growth rate of $64$. 
The generator (Figure~\ref{genblock}) gets as input a combination of randomly sampled noise and a multiple class count vector.
The concatenated vectors are passed through a fully connected layer (FC) with ReLU activation which is followed by the dense blocks.
The two dense blocks are coupled with a $1\times 1$ convolution to decrease the number of output feature maps and to improve computational efficiency and an upsampling layer to increase the spatial resolution.
The output feature maps from the dense block layers are forwarded to two $3\times 3$ convolutional layers (Conv) to generate images. 
For the discriminator as well as for the count network, we use four convolutional layers with shared weights followed by a fully connected layer to discriminate between real and fake images (discriminator) or to regress the multiple class count vector (count network). \\\\
\noindent\textbf{Implementation Details} The models are trained with images of size $64\times 64$ for Multi-MNIST and SVHN and $128\times 128$ for CLEVR images.
All images were scaled at the input with pixel values ranging from $-1$ to $1$.
Adam optimizer is used with momentum weights, $\beta_1 = 0.5$ and $\beta_2= 0.999$ respectively.
For generator and discriminator learning rate is fixed to $1e-4$.
The network is trained for $200$ epochs with batch size $128$ and count loss co-efficient $\lambda$ as 0.7. 
\subsection{MC$^2$-StyleGAN2}
We extended the official StyleGAN2 TensorFlow implementation corresponding to configuration-e for our count based image generation for CLEVR and SVHN images. 
Since the number of training images for CityCount is low, adaptive discriminator augmentation was applied while training the networks for CityCount images.
The mapping network is concatenated with the multiple-class count vector at each layers.
We also introduce dense like connections in place of residual connections in the synthesis/generator network for improved results.
We calculate the FID values on five samples of 50k generated images and report the average value. 
\\\\
\noindent\textbf{Implementation Details}  \\\\
\noindent\textbf{CLEVR and SVHN} The models are trained with images of size $64\times 64$ for SVHN and $128\times 128$ for CLEVR images.
The network is trained with minibatch size $32$ and the count loss co-efficient as $0.7$.
The rest of the hyperparameters is similar to the official implementation. \\\\
\noindent\textbf{CityCount} The models are trained with images of size $256\times 256$ for CityCount images.
For generator and discriminator, the learning rate is fixed to $0.0025$, and Adam optimizer is used with momentum weights, $\beta_1 = 0.0$ and $\beta_2= 0.99$ respectively.
The number of mapping layers used in the mapping network is $4$.
The rest of the hyperparameters is similar to the official implementation.
\subsection{Count prediction network}
\begin{figure*}[h]
         \centering
         \includegraphics[width=0.4\textwidth]{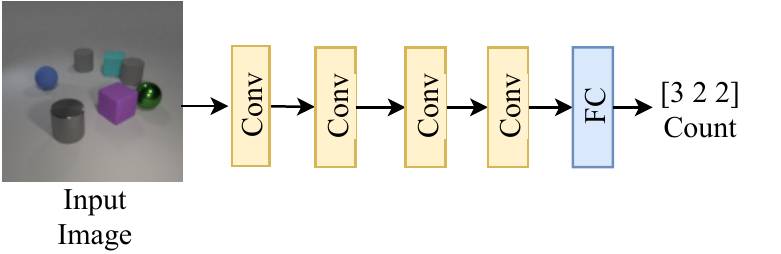}
    \caption{Count prediction network for CLEVR images. The network predicts the number of cylinders, spheres and cubes in the input RGB images as count vector}
    \label{countnw}
\end{figure*}

A convolution based network architecture shown in Figure~\ref{countnw} is used for the prediction of multiple-class count prediction of images.
The count regression network is similar to the one used in the count sub-network of the MC$^2$-SimpleGAN.
The network includes four convolution layers followed by LeakyRelu activation and dropout layers with the final block as a fully connected layer which outputs the multiple-class count vector of the corresponding input image.
\end{document}